\documentclass[10pt,twocolumns,journal,twoside,web]{IEEEtran}

\usepackage{textcomp}
\usepackage{amsmath}
\usepackage{amssymb}
\usepackage{amsfonts}
\usepackage{amsthm}
\usepackage{graphics}
\usepackage{multirow}
\usepackage{multicol}
\usepackage{epstopdf}
\usepackage{lineno}
\usepackage{commath}
\usepackage{graphicx}
\usepackage[justification=centering]{caption}
\usepackage{cite}
\usepackage{tabularx}
\usepackage{colortbl}
\usepackage{hhline}
\usepackage{subcaption}
\usepackage{comment}
\usepackage[colorlinks=true, allcolors=blue]{hyperref}
\usepackage{algorithmic}

\usepackage[shortlabels]{enumitem}
\usepackage{adjustbox}

\newtheorem{theorem}{Theorem}

\newtheorem{proposition}{Proposition}

\newtheorem{definition}{Definition}
\newtheorem{remark}{Remark}

\newtheorem{example}{{Example}}
\newtheorem{problem}{Problem}
\newtheorem{assumption}{Assumption}

\usepackage[linesnumbered,vlined,ruled,boxed]{algorithm2e}
\SetKwRepeat{Do}{do}{while}

\def\BibTeX{{\rm B\kern-.05em{\sc i\kern-.025em b}\kern-.08em
    T\kern-.1667em\lower.7ex\hbox{E}\kern-.125emX}}
\makeatletter
\def\endthebibliography{
\def\@noitemerr{\@latex@warning{Empty `thebibliography' environment}}%
\endlist
}

\begin{document}
\title{Structural Integrality in Task Assignment and Path Finding via Total Unimodularity of \\ Petri Net Models}
\author{Ioana Hustiu, Roozbeh Abolpour, Marius Kloetzer and Cristian Mahulea, \IEEEmembership{Senior Member, IEEE}
\thanks{This work was supported in part by grant PID2024-159284NB-I00 funded by MCIN/AEI/10.13039/501100011033 and by the ``European Union NextGenerationEU/PRTR'' and by Grant ONR-N62909-24-1-2081 funded by Office of Naval Research Global, USA. Corresponding author: Cristian Mahulea.}
\thanks{Ioana Hustiu and Marius Kloetzer are with the Department of Automatic Control and Applied Informatics, Technical University “Gheorghe Asachi” of Iasi, 700050 Iasi, Romania (e-mail: ioana.hustiu@academic.tuiasi.ro; marius.kloezter@academic.tuiasi.ro).}
\thanks{Roozbeh Abolpour is with the Energy Information Networks and
Systems Group at the Technical University of Darmstadt, 64289 Darmstadt, Germany (e-mail: roozbeh.abolpour@eins.tu-darmstadt.de).}
\thanks{Cristian Mahulea is with the Arag\'on Institute for Engineering Research (I3A), University of Zaragoza, 50018 Zaragoza, Spain (email: cmahulea@unizar.es).}}

\def\b#1{\mathchoice{\hbox{\boldmath $\displaystyle #1$}}
    {\hbox{\boldmath $\textstyle #1$}}
    {\hbox{\boldmath $\scriptstyle #1$}}
    {\hbox{\boldmath $\scriptscriptstyle #1$}}}
\newcommand{\preset}[1]{\ensuremath{\,\!^\bullet{#1}}}
\newcommand{\postset}[1]{\ensuremath{{#1}^\bullet}}
\newcommand{\tuple}[1]{\ensuremath{\langle {#1} \rangle}}

\maketitle

\begin{abstract}
Task Assignment and Path Finding (TAPF) concerns computing collision-free motions for multiple robots while jointly selecting goal locations. In this paper, safety is enforced by requiring unit-capacity traversal between successive intermediate markings, yielding coordination strategies that are valid independently of any specific time interpretation. Existing optimization-based approaches typically rely on time-expanded network-flow models, which result in large mixed-integer programs and limited scalability. We instead develop a Petri net (PN)–based optimization framework that exploits structural properties of the motion model to improve computational efficiency without explicit time expansion.

When robot motion is modeled by strongly connected state-machine PNs, we show that, once the congestion level (equivalently, the synchronization depth) is fixed to an integer value, the resulting motion-planning constraint matrix is totally unimodular. Consequently, the corresponding LP relaxation admits integral optimal solutions for the motion variables. When the estimated congestion exceeds one, we introduce a synchronization-on-demand mechanism based on intermediate markings; for a fixed number of synchronization stages, the associated constraint matrices remain totally unimodular, thereby preserving integrality of the motion variables.

Finally, we extend TAPF to Boolean specifications over regions of interest and propose a two-stage LP/mixed-integer linear programming (MILP) scheme in which integrality is confined to task-selection variables. Simulations on large benchmarks demonstrate substantial scalability improvements over time-expanded optimization baselines.

\end{abstract}

\begin{IEEEkeywords}
Multi-agent systems, combinatorial optimization, Petri nets, total unimodularity, linear programming.
\end{IEEEkeywords}

\maketitle

\section{Introduction}

\IEEEPARstart{P}{ath} planning for mobile robots is a fundamental problem in multi-agent systems, arising in applications such as industrial automation, logistics, and autonomous exploration. Classical approaches are often based on graph-search techniques \cite{mathew2015planning, debord2018trajectory}, while more recent methods seek scalability by combining sequential planning with time-optimal or safety-constrained trajectory generation \cite{robinson2018efficient}, or by incorporating heuristic strategies into the planning process \cite{yu2016optimal, ardizzoni2024constrained}. As the number of agents increases, the need to coordinate multiple robots operating in a shared environment leads to significant computational and modeling challenges \cite{guo2016task}.

Mixed-integer linear programming (MILP) formulations provide a flexible and expressive framework for multi-robot path planning, allowing a unified treatment of motion constraints, collision avoidance, and task allocation \cite{song2016rolling, ghassemi2022multi, schouwenaars2001mixed}. Despite their modeling power, MILP-based approaches are inherently limited by their computational complexity. In practice, exact MILP solvers scale only to relatively small problem instances, as the number of robots, tasks, or environment size grows \cite{richards2002aircraft}. This limitation has motivated the development of alternative approaches, including heuristic algorithms \cite{song2016rolling}, suboptimal iterative methods \cite{kalyanam2018scalable}, and tailored decomposition techniques such as branch-and-Benders-cut schemes \cite{alfandari2022tailored}.

To address the complexity of multi-robot coordination problems, formal modeling frameworks such as Boolean logic, temporal logic, and Petri nets (PNs) have been increasingly adopted \cite{BOMaKlGo20}. In particular, PNs offer a natural representation of concurrency, synchronization, and resource sharing, making them well suited for modeling coordinated robot motion and task execution. They have been successfully combined with temporal logic specifications to generate safe trajectories with favorable computational properties \cite{hustiu2024multi}, and have been used to encode complex coordination and interaction rules in multi-robot systems \cite{ziparo2011petri}. These features make PN-based models an attractive foundation for the formal analysis and control of multi-robot systems \cite{lacerda2019petri}.

Motivated by this context, this paper addresses the task allocation and safe path planning problem for multi-robot systems with Boolean specifications on the final team configuration, using PN models as the underlying formalism. The proposed approach builds upon the integer linear programming (ILP) and MILP formulations introduced in \cite{ARMaKl18} and \cite{abolpour2024optimizing}, and focuses on exploiting structural properties of the resulting optimization problems.

\section{Related work and contributions}\label{sec:2}
Multi-Agent Path Finding (MAPF) concerns the computation of collision-free paths (i.e., safe paths under a discrete-time occupancy model) for multiple agents moving in a shared environment with fixed start–goal assignments. When goal locations are not preassigned and must be selected jointly with the paths, the problem is referred to as Task Assignment and Path Finding (TAPF) \cite{ma2016optimal}. Both problems have received significant attention due to their relevance in robotics, logistics, and autonomous systems \cite{sharon2015conflict, ma2019searching, liu2021integrated, li2022mapf}. The problem considered in this paper is closely related to TAPF, but further generalizes it by allowing the global objective to be specified as a Boolean formula over regions of interest, rather than as a fixed set of target locations.

\emph{MAPF and TAPF approaches}. A seminal result for the classical TAPF problem is presented in \cite{yu2013multi}, where TAPF is reduced to a network flow problem on a time-expanded graph under unit-capacity constraints and collision-avoidance assumptions. This reduction yields LP-solvability, but relies on full temporal discretization and enforces synchronization at every discrete time step, leading to large problem instances for dense environments or long horizons.

In contrast, the approach proposed in this paper avoids explicit time expansion. Synchronization is introduced only when necessary through intermediate markings in a PN model, allowing robots to move in parallel whenever possible. Moreover, while \cite{yu2013multi} addresses TAPF with fixed goal sets, the proposed framework extends the problem formulation to global Boolean specifications, enabling automatic task selection jointly with path planning.

\emph{Optimization based and PN-based methods}. Optimization based formulations, particularly those based on ILP or MILP problems, offer a flexible framework for encoding motion constraints, collision avoidance, and task allocation \cite{schouwenaars2001mixed, song2016rolling, ghassemi2022multi}. However, such formulations are NP-hard in general and scale poorly with the number of robots, motivating the use of heuristics, decomposition strategies, or suboptimal relaxations \cite{richards2002aircraft, alfandari2022tailored}.

PNs have been extensively used as formal models for multi-robot coordination, due to their ability to compactly represent concurrency, synchronization, and shared resources \cite{ziparo2011petri, lacerda2019petri}. Prior work has demonstrated that Petri nets can be combined with logical task specifications to generate collision-free plans \cite{MAHULEA2020101, hustiu2024multi}. However, existing PN-based optimization approaches typically rely on ILP or MILP formulations, which inherit the scalability limitations of integer programming.


\subsection*{Contributions of this paper}

This paper establishes a structural optimization framework for task assignment and path finding in multi-robot systems with Boolean goal specifications. The main contributions are:

1) \emph{Congestion-aware reformulation of TAPF via state-machine PNs}.
We show that classical TAPF can be interpreted within a state-machine Petri net framework as a flow-conservation problem equivalent to standard network-flow formulations when unit-capacity constraints are enforced. Unlike classical approaches, we introduce congestion as an explicit optimization variable, which allows the planner to trade off parallelism and coordination. When the optimal solution achieves a unitary value of the congestion variable, the resulting LP relaxation yields safe and optimal trajectories, consistent with classical flow-based TAPF results.
When the congestion value exceeds one, the formulation provides quantitative information that can be used to guide collision-resolution strategies. This establishes a direct structural connection between network-flow TAPF formulations and PN-based motion models, while extending them to congestion-aware optimization.

2) \emph{Synchronization-on-demand with preserved structural integrality}.
Building on the congestion-aware formulation, we introduce a synchronization-on-demand mechanism based on intermediate PN markings, which enforces collision avoidance only when required. For a fixed number of synchronization stages, we show that the extended constraint matrices remain totally unimodular.
As a result, LP relaxations yield integer firing vectors and markings at each stage, enabling the construction of safe robot trajectories without resorting to full time expansion. This approach allows maximal parallel motion between synchronization points and avoids the combinatorial growth associated with fully time-expanded formulations.

3) \emph{Extension to Boolean task specifications with confined integrality}.
We generalize TAPF by allowing the global objective to be expressed as a Boolean formula over regions of interest, enabling joint task selection and path planning.
We propose a two-stage LP/MILP framework in which integrality is confined to the Boolean task-selection variables, while motion-related variables retain their integrality due to structural properties of the PN formulation. This separation yields a scalable optimization scheme that supports rich logical task
specifications while preserving optimality with respect to the congestion-aware cost function.

\section{Preliminaries and problem definition} \label{sec:3}

\subsection{Problem setting and notation} 
Consider a team of \( n_R \) identical mobile robots operating in a known, static, and discretized environment partitioned into a finite set of cells
\( \mathcal{C} = \{c_1, \dots, c_{n_{\mathcal{C}}}\} \). The workspace contains a finite set of disjoint regions of interest
\( \mathcal{Y} = \{y_1, \dots, y_{n_{\mathcal{Y}}}\} \),
which are used to specify the mission of the robotic team. The association between cells and regions is defined by a labeling function
\( h : \mathcal{C} \rightarrow \mathcal{Y} \cup \{\emptyset\} \),
where \( h(c_i) = y_j \) indicates that cell \( c_i \) belongs to region \( y_j \), and
\( h(c_i) = \emptyset \) denotes a free-space cell not associated with any region of interest.
As the robots move through the environment, their positions determine which regions of interest are observed. In particular, if at least one robot occupies a cell labeled with region \( y_j \) and all other robots are located in free-space cells, the team is said to observe region \( y_j \) according to the labeling function \( h \).

The workspace is modeled as a directed graph $\mathcal{G} = (\mathcal{C}, \mathcal{E}),$ where each vertex \( c_i \in \mathcal{C} \) corresponds to a cell of the discretized environment, and each directed edge
\( (c_i, c_j) \in \mathcal{E} \) represents a feasible motion of a robot from cell \( c_i \) to an adjacent cell \( c_j \).
Multiple robots may occupy the same cell, subject to collision-avoidance constraints introduced later.

To model robot motion and coordination over the environment graph, we employ a
Robot Motion Petri Net (RMPN) \cite{BOMaKlGo20}.
The Petri net is constructed directly from the environment graph
\( \mathcal{G} = (\mathcal{C}, \mathcal{E}) \) as follows. Each cell \( c_i \in \mathcal{C} \) is associated with a place \( p_i \in P \),
and each edge \( (c_i,c_j) \in \mathcal{E} \) is associated with a
transition \( t_k \in T \), representing a possible robot movement from
cell \( c_i \) to cell \( c_j \).
The pre-incidence matrix
\( \b{Pre} \in \{0,1\}^{|P|\times|T|} \)
and the post-incidence matrix
\( \b{Post} \in \{0,1\}^{|P|\times|T|} \)
encode the adjacency of the environment graph.
In particular, for each edge \( (c_i,c_j) \in \mathcal{E} \), there exists
a transition \( t_k \in T \) such that
\( \b{Pre}[p_i,t_k]=1 \) and \( \b{Post}[p_j,t_k]=1 \), corresponding to an arc
from place \( p_i \) to transition \( t_k \) and an arc from transition \( t_k \)
to place \( p_j \).

The resulting PN is a \emph{strongly connected state machine}, since by construction each transition
has exactly one input place and one output place.
Robots are represented as indistinguishable tokens, and the initial distribution
of robots is captured by the initial marking
\( \b{m}_0 \in \mathbb{N}_{\geq 0}^{|P|} \), where
\( \b{m}_0[p_i] \) denotes the number of robots initially located in cell \( c_i \).

With a slight abuse of notation, we write \( h(p_i) \) to denote \( h(c_i) \) for
the place \( p_i \) corresponding to cell \( c_i \). The association between the places of the RMPN and the regions of interest
\( \mathcal{Y} \) is encoded by the matrix $\b{V} \in \{0,1\}^{n_{\mathcal{Y}} \times n_P},$ defined such that \( \b{V}[i,j] = 1 \) if \( h(p_j) = y_i \), and
\( \b{V}[i,j] = 0 \) otherwise. Given a marking \( \b{m} \), the vector
\( \b{V}\cdot \b{m} \in \mathbb{N}^{n_{\mathcal{Y}}} \) represents the current observation of the robotic team, where each component greater than zero indicates whether the corresponding region of interest is occupied by at least one robot. In particular, \( \b{V}\cdot \b{m} = \b{0} \) indicates that all robots are in free-space cells.

\begin{assumption}\label{ass:unique_region_cell}
Each region of interest $y_i\in\mathcal{Y}$ corresponds to at most one cell of the discretized environment.
Equivalently, for each row of the matrix $\b{V}$, there exists exactly one nonzero entry.
As a consequence, $\b{V}$ is a submatrix of the identity matrix $\b{I}^{n_P}$ (up to row reordering).
\end{assumption}

This assumption implies that region occupancy in the final marking (constraint (b) in MILP \eqref{eq:milp_imb}) can be expressed through simple bound constraints on individual places if $\b{x}$ is fixed, which is essential for preserving the total unimodularity of the motion-planning constraint matrices.

Let \( \b{C} = \b{Post} - \b{Pre} \) denote the incidence (token-flow) matrix of the PN. A transition \( t_j \in T \) is enabled at a marking \( \b{m} \) if its unique input place \( \preset{t_j} = \{p\in P | \b{Pre}[p,t_j]=1\}\) contains at least one token, that is, \( \b{m}[\preset{t_j}] \geq 1 \). Firing an enabled transition \( t_j \) consumes one token from its input place \( \preset{t_j} \) and produces one token at its output place \( \postset{t_j} = \{p \in P | \b{Post}[p,t_j]=1\} \), thereby representing the motion of a robot between adjacent
cells.

Given a firing count vector \( \b{\sigma} \in \mathbb{N}^{|T|}_{\geq 0} \), where
\( \b{\sigma}[j] \) denotes the number of times transition \( t_j \) fires, the
resulting marking \( \b{m}_f \) is determined from \( \b{m}_0 \) by the PN state equation
\begin{equation}\label{eq:fund}
\b{m}_f = \b{m}_0 + \b{C}\cdot \b{\sigma}.
\end{equation}

\begin{remark}
For strongly connected state machine Petri nets, the state equation \eqref{eq:fund} is a necessary and sufficient condition for reachability of $\b{m}_f$ \cite{ICSiTeCo98}. Moreover, when $\b{\sigma}$ is obtained as the solution of a
minimum total firing optimization problem, an explicit firing sequence realizing $\b{\sigma}$ can be constructed efficiently (see~\cite{ARMaKl18} for a constructive algorithm) since $\b{\sigma}$ cannot contain empty cycles.
\end{remark}

\begin{figure}[ht]
    \centering
    \captionsetup{justification=centering}
    \includegraphics[width=\columnwidth]{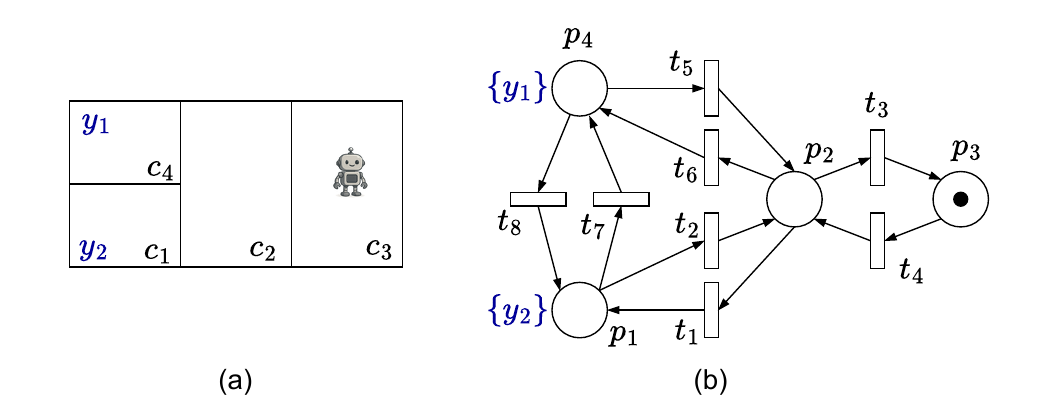}
\caption{Example of a RMPN.
(a) Discretized environment with four cells and two regions of interest.
(b) Corresponding RMPN encoding feasible robot motions.}
    \label{fig:RMPN}
\end{figure}

\begin{example}\label{ex:RMPN}
Consider the discretized environment shown in Fig.~\ref{fig:RMPN}(a), consisting of four cells
\( \{c_1, c_2, c_3, c_4\} \) and two regions of interest
\( \mathcal{Y} = \{y_1, y_2\} \).
Cells \( c_1 \) and \( c_4 \) are associated with regions \( y_2 \) and \( y_1 \), respectively,
while cells \( c_2 \) and \( c_3 \) correspond to free space.
A single mobile robot is initially located in cell \( c_3 \).
The environment is modeled as a directed graph whose edges represent feasible robot motions
between adjacent cells.

Fig.~\ref{fig:RMPN}(b) illustrates the corresponding RMPN that describes this environment.
The net consists of the set of places
\( P = \{p_1, p_2, p_3, p_4\} \),
the set of transitions
\( T = \{t_1, \dotsc, t_8\} \),
and the initial marking
\( \b{m}_0 = [0,\,0,\,1,\,0]^\top \).
The incidence matrix of the RMPN is given by
\[
\begin{array}{l}
\ \ \ \ \ \ \ \ \ \ \ \ \ \\[-1mm]
\ \ \ \ \ \ \ \ \ \ \ \ \ \ \ \ \ \ t_1 \ \ \ \ t_2 \ \ \ \ t_3 \ \ \ \ t_4 \ \ \ \ t_5 \ \ \ \ t_6 \ \ \ \ t_7 \ \ \ \ t_8 \\
\b{C}
=
\begin{array}{r}
p_1 \\ p_2 \\ p_3 \\ p_4
\end{array}
\left[
\begin{array}{rrrrrrrr}
1 & -1 & 0 & 0 & 0 & 0 & -1 & 1 \\
-1 & 1 & -1 & 1 & 1 & -1 & 0 & 0 \\
0 & 0 & 1 & -1 & 0 & 0 & 0 & 0 \\
0 & 0 & 0 & 0 & -1 & 1 & 1 & -1
\end{array}
\right].
\end{array}
\]
Each column of the incidence matrix \( \b{C} \) represents the effect of firing the
corresponding transition.
For instance, transition \( t_1 \) models the movement of the robot from cell \( p_2 \)
to cell \( p_1 \).

The labeling function satisfies
\( h(p_1)=\{y_2\} \), \( h(p_2)=\emptyset \),
\( h(p_3)=\emptyset \), and \( h(p_4)=\{y_1\} \).
Accordingly, the characteristic matrix is
\[
\begin{array}{l}
\ \ \ \ \ \ \ \ \ \ \ \ \ \ \ p_1 \ \ p_2 \ \ p_3 \  p_4 \\
\b{V}
=
\begin{array}{r}
y_1 \\ y_2
\end{array}
\left[
\begin{array}{rrrr}
0 & 0 & 0 & 1 \\
1 & 0 & 0 & 0
\end{array}
\right].
\end{array}
\]
Since the initial marking is \( \b{m}_0 \), we obtain
\( \b{V}\cdot \b{m}_0 = [0,\,0]^\top \), meaning that in its initial position the robot
does not observe any region of interest, as it is located in the free-space cell \( p_3 \).

Assume that transitions \( t_4 \) and \( t_1 \) are fired.
The corresponding firing count vector is
\( \b{\sigma} = [1,\,0,\,0,\,1,\,0,\,0,\,0,\,0]^\top \).
According to~\eqref{eq:fund}, the final marking is
\( \b{m}_f = [1,\,0,\,0,\,0]^\top \).
Computing \( \b{V}\cdot \b{m}_f \) yields \( [0,\,1]^\top \), which indicates that the robot
is observing the region labeled \( y_2 \), since \( h(p_1)=\{y_2\} \). \hfill $\blacksquare$
\end{example}

\begin{problem}\label{pb:1}
Given a team of \( n_R \) mobile robots operating in a discretized environment, an associated RMPN with initial marking \( \b{m}_0 \), and a Boolean formula \( \varphi \) defined over the set of regions of interest \( \mathcal{Y} \), determine a task allocation and a set of safe robot motions such that the resulting final marking \( \b{m}_f \) is reachable from \( \b{m}_0 \) and satisfies the global Boolean specification \( \varphi \). 
\end{problem}

Problem~\ref{pb:1} extends the classical TAPF formulation by imposing a Boolean
specification on the final configuration of the robotic team. A
trajectory is defined as the sequence of PN transitions executed by a
robot from its initial position to its destination.

The Boolean formula $\varphi$ encodes a global task by specifying regions to be
visited or avoided in the final marking. Only final-state specifications are
considered, and thus we exclude cases where more than \( n_R \) regions must be visited. Without loss of generality, $\varphi$ is assumed to be in
conjunctive normal form (CNF),
$\varphi=\varphi_1\wedge\dots\wedge\varphi_{n_d}$, where each clause $\varphi_i$
is a disjunction of literals over $\mathcal{Y}$. Following~\cite{ARMaKl18}, the
Boolean specification is translated into a system of linear inequalities.

To represent region visitation in the final marking, we introduce a binary
vector $\b{x}\in\{0,1\}^{n_\mathcal{Y}}$, where $x_i=1$ if region $y_i$ is
visited. The Boolean constraint $\varphi$ is then encoded as
$\b{A}_{\varphi}\cdot \b{x}\le\b{b}_{\varphi}$, where $\b{A}_{\varphi}\in\{-1,0,1\}^{n_d\times n_\mathcal{Y}}$ and $\b{b}_{\varphi}\in\mathbb{N}_{\ge0}^{n_d}$ are constructed exactly as in \cite{ARMaKl18}.

\textbf{Safety via unit-capacity traversal.} To guarantee safe execution independent of any specific time movement, we impose that each cell $p_i$ is traversed by at most one robot when moving from the initial marking $\b{m}_0$ to the final marking $\b{m}_f$. This strict condition can be written straightforwardly:

\begin{equation}\label{eq:s_equal_1}
    \b{Post} \cdot \b{\sigma} + \b{m}_0 \leq \b{1}.
\end{equation}

This condition guarantees parallel free movements (without synchronization) from $\b{m}_0$ to a final marking $\b{m}_f$. 

In many practical cases, however, condition \eqref{eq:s_equal_1} may be too restrictive. For example, when a narrow passage, e.g., a bridge, must be crossed by several robots to reach their destinations. To bound congestion during execution, we impose:

\begin{equation}\label{eq:s_var}
    \b{Post} \cdot \b{\sigma} + \b{m}_0 \leq s \cdot \b{1}.
\end{equation}

\begin{remark}
Constraint~\eqref{eq:s_var} does not model simultaneous occupancy in physical time. Instead, it provides a conservative bound on the maximum number of tokens that may traverse a place over the execution represented by $\b{\sigma}$. Its role is therefore to act as a congestion surrogate whose value guides the introduction of synchronization stages, rather than as a direct temporal collision-avoidance constraint.
\end{remark}

Safe execution is obtained only when $s=1$, or after introducing
synchronization stages when $s>1$.

When the variable $s$ is penalized with a sufficiently large weight in the cost function, the optimal solution of an optimization problem subject to~\eqref{eq:s_var}, satisfies $s \;=\; \left\lVert \b{Post}\cdot \b{\sigma} + \b{m}_0 \right\rVert_\infty,$ i.e., $s$ equals the maximum number of tokens generated to any place over the firing count vector solution from $\b{m}_0$ to $\b{m}_f$. Therefore, minimizing $s$ enforces a solution with minimal congestion, and the remaining objective term selects, among all minimum-congestion solutions, one with minimal total number of transition firings.

If the optimal solution\footnote{Throughout the paper, the symbol `*' indicates the optimal solution.} after solving an optimization problem yields $s^* = 1$, then fully parallel, collision-free trajectories are obtained directly. Otherwise, if $s^*>1$, we will introduce $\lceil s^* \rceil$ intermediate markings, which act as synchronization points ensuring that at most one enters a region at a time.  For example, having $s^* = 2$ means that starting from the initial marking $\b{m}_0$, (at least) 2 intermediary markings (the second one being the final marking $\b{m}_f$) will be necessary to obtain safe paths. Unlike the solution in \cite{MAHULEA2020101}, which relies on assigning priorities, in this paper we explicitly construct such intermediate markings. In some situations, introducing exactly $\lceil s^* \rceil$ synchronizations may still be insufficient due to multiple interconnected narrow passages. Feasibility can usually be obtained by incrementing the number of intermediate markings by one. However, this number is upper-bounded by the number of robots $n_R$, case in which the robots will  move sequentially one after the other to their final destinations. In our simulations, some of them with more than 2.000 robots, one or two such increments were always sufficient to obtain a feasible, collision-free solution. Between two consecutive synchronizations, robots again move in parallel.

The complete robot path planning problem for the extended TAPF can now be formulated as follows \cite{ARMaKl18}:

\begin{equation}\label{eq:milp}
\begin{aligned}
\text{Minimize} \quad & \b{1}^T \cdot \b{\sigma} + M \cdot s \\
\text{Subject to:} \quad & \b{m} = \b{m}_0 + \b{C} \cdot \b{\sigma}, & (a)\\
& \b{x} \leq \b{V} \cdot \b{m} \leq N \cdot \b{x}, & (b)\\
& \b{A}_{\varphi} \cdot \b{x} \leq \b{b}_{\varphi}, & (c)\\
& \b{Post} \cdot \b{\sigma} + \b{m}_0 \leq s \cdot \b{1}^{n_P}, & (d)\\
& \begin{array}{l}
\b{m} \geq \b{0}^{n_P}, \b{\sigma} \geq \b{0}^{|T|},  \\
\b{0}^{n_\mathcal{Y}} \leq \b{x} \leq \b{1}^{n_\mathcal{Y}},  s \geq 1.
\end{array} & (e)\\
\end{aligned}
\end{equation}

In \eqref{eq:milp}, the unknowns are represented by the tuple $(\b{m}, \b{\sigma}, \b{x}, s)$, while $N > n_R + 1$ and $M > |T|+1$ are sufficiently large positive constants used to force feasibility and prioritize collision avoidance. We will report only the first term of the cost function, as the second term serves only as an auxiliary factor in the optimization and not as a performance metric. The constraints have the following meaning: $(a)$ is the state equation \eqref{eq:fund}, linking the initial marking $\b{m}_0$, the final marking $\b{m}$, the incidence matrix $\b{C}$, and the firing vector $\b{\sigma}$; $(b)$ relates the final marking $\b{m}$ with the Boolean variables $\b{x}$, using the labelling matrix $\b{V}$; $(c)$ encodes the global Boolean specification on the final state; $(d)$ together with the second term in the cost function corresponds to the minimization of $s=||\b{Post} \cdot \b{\sigma} + \b{m}_0||_{\infty}$. This variable $s$ captures the maximum congestion along the path and $(e)$ enforces boundary conditions for the unknowns.

\begin{remark}
Since our approach relies on LP relaxations, we will denote the problem type as \eqref{eq:milp} – $(\cdot)$. Specifically, \eqref{eq:milp} – ILP uses only integer variables, \eqref{eq:milp} – MILP treats most of the variables as continuous while keeping a subset as integer, and \eqref{eq:milp} – LP relaxes all variables to be continuous.
\end{remark}

\subsection{Modeling assumptions and scope}
The proposed framework is developed under a set of modeling assumptions that delimit its scope and applicability. First, the environment is represented as a discretized workspace, and robot motion is restricted to transitions between adjacent cells, which are captured by a state-machine Petri net. At any given marking, each cell can be occupied by at most one robot, and collisions are prevented by enforcing mutual exclusion constraints on conflicting places. This abstraction is standard in TAPF and MAPF formulations and enables a compact representation of collision avoidance.

Second, robots are assumed to be indistinguishable at the motion-planning level, in the sense that any robot can occupy any cell and execute any feasible transition. This assumption is consistent with classical TAPF formulations and is essential for the total unimodularity properties established in this paper. Extensions to heterogeneous robots or robot-specific constraints would generally break the structural conditions required for integrality and are therefore outside the scope of this work.

Third, task specifications are imposed on the final marking of the Petri net and are expressed as Boolean formulas over regions of interest. Intermediate-state logical constraints and temporal logic specifications are not considered. While this restricts the expressiveness of the task layer, it allows integrality to be confined to a small set of task-selection variables, preserving scalability of the overall optimization framework.

Finally, collision avoidance is enforced through synchronization-on-demand using intermediate markings, which assumes that conflicts can be detected and resolved at discrete points. This mechanism avoids full time expansion but does not model continuous-time dynamics or uncertainty in robot motion. Addressing continuous-time effects, robustness, or dynamic environments is left for future work.

\section{Structural integrality via TU}\label{sec:TU}

The formulations in Sections~\ref{sec:4}--\ref{sec:solution} are derived from PN flow constraints and, in general, lead to integer or mixed-integer programs. The key observation exploited in this paper is that \emph{integrality of the motion variables can be recovered structurally} from total unimodularity (TU) once certain parameters (e.g., a congestion bound or the number of synchronization stages) are fixed to integer values. This section recalls some TU facts relevant to our problem.

\begin{definition}\label{def:TU}
A matrix $\b{A}\in\mathbb{Z}^{m\times n}$ is \emph{totally unimodular (TU)} if every square submatrix of $\b{A}$ has determinant $0$ or $\pm 1$.
\end{definition}

The relevance of TU is that it yields integrality of LP vertices under integer right-hand sides.

\begin{theorem}[TU implies integrality of LP vertices {\cite{BOSchrijver98}}]\label{th:TU_LP}
Let $\b{A}\in\mathbb{Z}^{m\times n}$ be TU and let $\b{b}\in\mathbb{Z}^{m}$.
Then every vertex of the polyhedron
\[
\{\b{z}\in\mathbb{R}^{n}_{\ge 0}\mid \b{A}\b{z}\le \b{b}\}
\]
is integral. In particular, any feasible LP of the form $\min \b{c}^\top \b{z}$ subject to $\b{A}\b{z}\le \b{b}$ and $\b{z}\ge 0$ admits an integral optimal solution.
\end{theorem}

To establish TU for the constraint matrices arising from the Petri net motion model, we use the following characterization.

\begin{theorem}[Ghouila--Houri criterion {\cite{BOSchrijver98}}]\label{th:GH}
A matrix $\b{A}\in\mathbb{R}^{m\times n}$ is TU if and only if, for every subset of rows $R\subseteq\{1,\dots,m\}$, there exists a partition $R=R_1\cup R_2$ such that for each column $j$,
\[
\sum_{i\in R_1} A_{ij}-\sum_{i\in R_2} A_{ij}\in\{-1,0,1\}.
\]
\end{theorem}

\begin{remark}[How TU is used in this paper]\label{rem:conditional_TU}
The overall task-assignment-and-path-finding problem remains mixed-integer in general because of the Boolean task-selection variables.
TU is used \emph{conditionally} to recover integrality of the motion variables:
\begin{itemize}
\item in the TAPF subproblem, $\b{\sigma}$ becomes integral in the LP relaxation once the congestion bound $s$ is fixed to an integer value (Section~\ref{sec:4});
\item in the synchronization-on-demand refinement, all intermediate firing vectors and markings become integral in the LP relaxation once the number of synchronization stages $\bar{s}$ is fixed (Section~\ref{sec:4}).
\end{itemize}
Accordingly, integrality is \emph{not} claimed when $s$ (or $\bar{s}$) is a continuous decision variable or when Boolean variables remain relaxed.
\end{remark}

The next section applies Theorem~\ref{th:GH} to the PN motion constraints and establishes total unimodularity of the relevant matrices under the above fixed-parameter conditions.

\section{Task-Assignment and Path Finding Problem} \label{sec:4}
This section isolates the structural core of the proposed approach.
We first show that the classical TAPF problem arises as a special case of Boolean specifications, then clarify why congestion optimization breaks standard flow interpretations, and finally establish conditional integrality results that support the scalability of the method.

\subsection{TAPF as a special case of Boolean specifications}

We consider first the classical TAPF setting, in which a team of $n_R$ robots must be assigned to a set of $n_{\mathcal{Y}}=n_R$ disjoint regions of interest, with each region required to be occupied by exactly one robot in the final configuration.
This corresponds to the Boolean specification
\[
\varphi = y_1 \wedge y_2 \wedge \dots \wedge y_{n_{\mathcal{Y}}}.
\]

Under this specification, the Boolean decision vector is fixed to $\b{x}=\b{1}^{n_\mathcal{Y}}$, and the final marking is uniquely determined as
\[
\b{m}_f = \b{V}^\top \cdot \b{x}.
\]
Consequently, task selection $\b{x}$ is no longer a decision variable and the problem reduces to computing collision-free robot motions that transfer the initial marking $\b{m}_0$ to $\b{m}_f$ while minimizing the total number of firings.

Substituting this specification into the general formulation~\eqref{eq:milp}, we obtain the reduced optimization problem
\begin{equation}\label{eq:milp_m}
\begin{aligned}
\text{Minimize} \quad & \b{1}^\top \cdot \b{\sigma} + M \cdot s \\
\text{Subject to} \quad 
& \b{C}\cdot \b{\sigma} = \b{m}_f - \b{m}_0, \\
& \b{Post}\cdot \b{\sigma} + \b{m}_0 \le s\cdot \b{1}^{n_P}, \\
& \b{\sigma} \ge \b{0}^{|T|}, \quad s \ge 1.
\end{aligned}
\end{equation}

The decision variables in~\eqref{eq:milp_m} are the firing-count vector $\boldsymbol{\sigma}$ and the scalar $s$, which represents a congestion bound.
The objective minimizes the total number of transition firings, while penalizing congestion through the auxiliary variable $s$.

\subsection{Relation to classical network flow}

At first glance, problem~\eqref{eq:milp_m} resembles a minimum-cost flow problem.
Indeed, when robot motion is modeled by a state-machine Petri net, the incidence matrix $\b{C}$ plays a role analogous to the adjacency matrix of a directed graph, and the constraint $\b{C}\cdot \b{\sigma}=\b{m}_f-\b{m}_0$ enforces flow conservation.

However, there is a fundamental difference with respect to classical network-flow formulations.
In standard minimum-cost flow problems, arc/node capacities are fixed parameters, and integrality of the LP relaxation follows directly from the total unimodularity of the incidence matrix.
In contrast, in~\eqref{eq:milp_m} the congestion bound $s$ is itself a decision variable.
The inequality $\b{Post}\cdot \b{\sigma}+\b{m}_0 \le s\cdot \b{1}$ does not impose a fixed capacity, but introduces a \emph{conservative congestion surrogate} whose value is optimized jointly with the motion plan.

As a result, problem~\eqref{eq:milp_m} cannot be interpreted as a single capacitated flow problem.
Instead, it should be viewed as a parametric family of flow problems indexed by the congestion level $s$.
For a given fixed integer value of $s$, the feasible set of~\eqref{eq:milp_m} coincides structurally with that of a classical capacitated network flow problem.
The role of the optimization over $s$ is to determine the smallest congestion level for which a feasible flow exists.

\begin{remark}
It is important to emphasize that the novelty of this result does not lie in the integrality of network flows per se, which is classical, but in identifying a congestion–parametrized family of flow–like formulations arising from Petri–net motion models.
Unlike time-expanded formulations, where capacities are fixed a priori, here integrality is recovered only after conditioning on the congestion level or synchronization depth.
This conditional integrality property support the proposed synchronization-on-demand mechanism and has no direct analogue in standard TAPF network-flow models.
\end{remark}

\subsection{Total unimodularity for fixed congestion}

The following result formalizes the above observation and establishes the integrality properties required for scalability.

\begin{theorem}\label{th:TU_final_m}
Consider a state-machine Petri net and fix an integer value of the congestion bound $s$.
Then the constraint matrix associated with the LP obtained from~\eqref{eq:milp_m} by fixing $s$,
namely
\[
\begin{bmatrix}
\b{C} \\
\b{Post}
\end{bmatrix},
\]
is totally unimodular. Consequently, the corresponding LP admits an optimal solution with integer firing vector $\b{\sigma}$.
\end{theorem}

\begin{proof}
The proof is provided in Appendix~\ref{ap:TU1}.
\end{proof}

Theorem~\ref{th:TU_final_m} implies that, for any fixed integer $s$, the LP relaxation of~\eqref{eq:milp_m} admits an optimal solution in which the firing vector $\boldsymbol{\sigma}$ is integer.
This result is structurally equivalent to the classical integrality property of minimum-cost flow problems, but it is stated here explicitly because it forms the basis for the extensions developed next.

It is important to emphasize that integrality is not guaranteed for~\eqref{eq:milp_m} when $s$ is treated as a continuous decision variable.
Instead, integrality of $\boldsymbol{\sigma}$ follows once $s$ is fixed to an integer value.
This distinction is crucial for understanding the role of congestion optimization in the proposed framework.

\subsection{From congestion optimization to synchronization}

After solving the LP relaxation of~\eqref{eq:milp_m}, two cases arise.
If the optimal solution satisfies $s^*=1$, then the resulting firing vector $\boldsymbol{\sigma}^*$ corresponds to a fully parallel, collision-free plan and is guaranteed to be integer by Theorem~\ref{th:TU_final_m}.
No further refinement is required.

If instead $s^*>1$, collisions may occur if the plan encoded by $\boldsymbol{\sigma}^*$ is executed without coordination.
In this case, the value $s^*$ is interpreted as an estimate of the minimum number of synchronization phases required to resolve congestion.
We therefore introduce $\bar{s}=\lceil s^* \rceil$ intermediate markings that partition the motion into $\bar{s}$ stages, each enforcing unit-capacity constraints.

This idea leads to the following multi-stage formulation, where the cost function prioritizes movements in early stages:
\begin{equation}\label{eq:milp_mc}
\begin{aligned}
\text{Minimize} \quad & 
   \sum_{i=1}^{\bar{s}} \b{1}^\top \cdot \b{\sigma}_i \cdot \bigl(1 + (\bar{s}-i)\bigr) \\
\text{Subject to} \quad 
& \b{m}_i = \b{m}_{i-1} + \b{C}\cdot \b{\sigma}_i, \quad i=1,\dots,\bar{s}, \\
& \b{Post}\cdot \b{\sigma}_i + \b{m}_{i-1} \le \b{1}^{n_P}, \quad i=1,\dots,\bar{s}, \\
& \b{m}_{\bar{s}} = \b{m}_f, \\
& \b{m}_i \ge \b{0}^{n_P}, \quad \b{\sigma}_i \ge \b{0}^{|T|}
\end{aligned}
\end{equation}

Here, $\b{m}_i$ and $\boldsymbol{\sigma}_i$ denote the intermediate markings and firing vectors at each synchronization step.
Between two consecutive markings, unit-capacity constraints are enforced explicitly, guaranteeing safe execution.

\subsection{Preservation of integrality with intermediate markings}

The introduction of intermediate markings does not destroy the integrality properties of the formulation.

\begin{theorem}\label{th:inter}
For a state-machine Petri net, the constraint matrix associated with~\eqref{eq:milp_mc} is totally unimodular.
\end{theorem}

\begin{proof}
The proof is provided in Appendix~\ref{ap:TU2}.
\end{proof}

By Theorem~\ref{th:inter}, the LP relaxation of~\eqref{eq:milp_mc} admits an integral optimal solution for all firing vectors and intermediate markings.
Thus, collision-free robot trajectories can be computed using linear programming alone, even when multiple synchronization phases are required.

This result is essential for the remainder of the paper, as it shows that the classical integrality properties of network flows extend naturally to the synchronization-on-demand mechanism employed here.
In the next section, this structural insight is leveraged to handle general Boolean specifications while confining integrality to a small set of task-selection variables.

\section{Extended TAPF solution strategy}\label{sec:solution}

We now return to the complete formulation \eqref{eq:milp}, which includes the decision variable $\b{x}$.  
The solution proceeds in two stages that combine the LP relaxation ideas of Section~\ref{sec:TU} with the collision-avoidance mechanism described previously.  

\textbf{Stage~I: LP relaxation of \eqref{eq:milp}}.  
We first solve \eqref{eq:milp}--LP, where all variables $(\b{m},\b{\sigma},\b{x},s)$ are relaxed to be continuous.  

\begin{itemize}
\item If the relaxation is infeasible, then the original ILP is also infeasible, since the LP enlarges the feasible region.  
\item Otherwise, let $(\b{m}^*,\b{\sigma}^*,\b{x}^*,s^*)$ be the optimal LP solution.  
\end{itemize}

Two situations arise:  
\begin{enumerate}
\item If $s^*=1$ and $\b{x}^*$ is integer, the solution is feasible and safe.  
In this case, $\b{m}^*$ and $\b{\sigma}^*$ are guaranteed to be integer.  
Indeed, once $\b{x}^*$ is integers, constraint~\eqref{eq:milp}-(c) becomes redundant and \eqref{eq:milp}-(b) reduces to simple bounds on $\b{m}$ (since, by Assumption~\ref{ass:unique_region_cell}, $\b{V}$ is a submatrix of the identity).  
The remaining constraint matrix is
\[
\begin{bmatrix}
\b{C} & -\b{I} \\
\b{Post} & \b{0}
\end{bmatrix}.
\]  
From Theorem~\ref{th:TU_final_m}, we know that the vertical concatenation $\begin{bmatrix}\b{C} \\ \b{Post}\end{bmatrix}$ is TU.  
Appending the block $\begin{bmatrix}-\b{I} \\ \b{0}\end{bmatrix}$ corresponds to concatenating this TU matrix with an identity matrix, an operation that preserves total unimodularity (see, e.g., \cite{BOSchrijver98}).  
Therefore, the full matrix above is also TU.

\item Otherwise (i.e., if $s^*>1$ or $\b{x}^*$ is fractional), we set the number of required synchronizations to $\bar{s}=\lceil s^*\rceil$ and proceed to Stage~II.  
\end{enumerate}

\textbf{Stage~II: Safe refinement with intermediate markings}.  
In this stage, $\b{x}$ is explicitly kept as a binary decision variable, ensuring that the Boolean specification is satisfied exactly.  
We then construct the extended formulation \eqref{eq:milp_imb}, which introduces $\bar{s}$ intermediate markings to guarantee that at most one robot enters any cell between two consecutive synchronizations:  

\begin{equation}\label{eq:milp_imb}
\begin{aligned}
\text{Minimize} \quad & 
   \sum_{i=1}^{\bar{s}} \b{1}^T \cdot \b{\sigma}_i \cdot \left( 1 + (\bar{s}-i) \right) \\[0.1em]
\text{Subject to:} \quad 
& \b{m}_i = \b{m}_{i-1} + \b{C}\cdot \b{\sigma}_i, \quad i=1,\ldots,\bar{s}, & (a) \\
& \b{x} \leq \b{V}\cdot \b{m}_{\bar{s}} \leq N\cdot \b{x}, & (b) \\
& \b{A}_{\varphi}\cdot \b{x} \leq \b{b}_{\varphi}, & (c) \\
& \b{Post}\cdot \b{\sigma}_i + \b{m}_{i-1} \leq \b{1}^{n_P}, \quad i=1,\ldots, \bar{s}, & (d) \\
& \begin{array}{l}
    \b{m}_i \in \mathbb{R}_{\geq 0}^{|P|}, \ \b{\sigma}_i \in \mathbb{R}_{\geq 0}^{|T|},\; i=1,\ldots,\bar{s} \\
    \b{x} \in \{0,1\}^{n_{\mathcal{Y}}}
    \end{array}. & (e)
\end{aligned}
\end{equation}

Here, $\b{m}_i$ and $\b{\sigma}_i$ denote the markings and firing vectors at each synchronization step, while $\b{x}$ enforces the Boolean specification at the final state.  
By Theorem~\ref{th:inter}, the constraint matrix of \eqref{eq:milp_imb} (excluding the Boolean rows) is TU; hence, apart from the binary variables in $\b{x}$, MILP \eqref{eq:milp_imb} yields integer solutions for $\b{m}_i$ and $\b{\sigma}_i$.  
If the problem is infeasible, $\bar{s}$ can be incremented and the problem re-solved.  
In practice, we observed that only a few additional synchronizations were sufficient to restore feasibility. Moreover, the upper bound on the number of intermediate markings is equal to the number of robots $n_R$ since the robots can follow their trajectories sequentially.

\begin{algorithm}[t]
\small
\DontPrintSemicolon
\SetAlgoNlRelativeSize{-1}   
\SetAlgoNlRelativeSize{-1}
\LinesNumbered
\SetAlgoLined
\caption{Two–Stage Planner with Boolean Specs}
\label{alg:two_stage}
\KwIn{
RMPN $\Sigma_{\mathcal N}=\langle P,T,\mathbf{Pre},\mathbf{Post}\rangle$, initial marking $\mathbf{m}_0$; labeling matrix $\mathbf{V}$; Boolean constraints $\langle \mathbf{A}_\varphi,\mathbf{b}_\varphi \rangle$; big constants $M,N$.
}
\KwOut{
Safe plan: integer $\{\mathbf{m}_i,\b{\sigma}_i\}_{i=1}^{\bar{s}}$ and binary $\b{x}$; or \emph{infeasible}.
}

\BlankLine
\textbf{Stage I: Solve \eqref{eq:milp}--LP (all variables relaxed)}\;
Solve \eqref{eq:milp}--\textbf{LP} with decision variables $(\mathbf{m},\b{\sigma},\b{x},s)$\;
\If{\textbf{LP infeasible}}{
  \textbf{return} \emph{infeasible}\;
}
Let $(\b{m}^*,\b{\sigma}^*,\b{x}^*,s^*)$ be the optimal solution of \eqref{eq:milp}--\textbf{LP}\;

\BlankLine
\eIf{$s^*=1$ \textbf{and} $\b{x}^*\in\{0,1\}^{n_{\mathcal{Y}}}$}{
\tcp{Given integer $\b{x}^*$ and $s^*=1$, TU implies $\b{m}^*,\b{\sigma}^*$ are integer}
  \textbf{return} $\bar{s}=1$, $\b{m}_1=\b{m}^*$, $\b{\sigma}_1=\b{\sigma}^*$, $\b{x}=\b{x}^*$\;
}{
  $\bar{s}\leftarrow \lceil s^* \rceil$ \tcp*{min. \# synchronizations}
}

\BlankLine
\textbf{Stage II: Safe refinement with intermediate markings}\;
\For{$k := \bar{s}$ \KwTo $n_R$}{
  Solve \eqref{eq:milp_imb}--\textbf{MILP} with variables
  $\{\mathbf{m}_i,\b{\sigma}_i\}_{i=1}^{k} \in \mathbb{R}$,
  $\b{x}\in\{0,1\}^{n_{\mathcal{Y}}}$\;
  \If{\eqref{eq:milp_imb}--\textbf{MILP} feasible}{
     \tcp{By TU (Thm.~\ref{th:inter}), $\mathbf{m}_i,\b{\sigma}_i$ are integer}
     \textbf{return} $\bar{s} = k, \{\mathbf{m}_i,\b{\sigma}_i\}_{i=1}^{k}$, $\b{x}$\;
  }
}
\textbf{return} \emph{infeasible}\;
\end{algorithm}

\begin{proposition}\label{prop:completeness}
Consider Problem~\ref{pb:1} under the staged execution model induced by
\eqref{eq:milp_imb}, i.e., plans are represented by a finite number of
synchronization stages with stage-wise unit-capacity constraints
\eqref{eq:milp_imb}-(d).
Then Algorithm~\ref{alg:two_stage} is:
\begin{enumerate}
\item \emph{Sound}: whenever it returns a plan
$\{\b{m}_i,\b{\sigma}_i\}_{i=1}^{\bar{s}}$ and $\b{x}$, the returned plan is feasible
for \eqref{eq:milp_imb} and therefore satisfies the Boolean specification $\varphi$
and the safety constraints.
\item \emph{Complete up to $n_R$ stages}: if there exists an integer $k\in\{1,\dots,n_R\}$
such that \eqref{eq:milp_imb} is feasible with $k$ stages, then the loop in Stage~II
will find a feasible solution for some $\bar{s}\le k$ and return it.
\end{enumerate}
\end{proposition}

\begin{proof}
Soundness follows directly from construction: Stage~I returns only the LP solution
when the termination condition is met, and Stage~II returns only after a feasible
solution of \eqref{eq:milp_imb} is found. In both cases, the returned variables satisfy
the corresponding constraints, hence $\varphi$ is satisfied by \eqref{eq:milp_imb}-(b)--(c)
and unit-capacity is enforced by \eqref{eq:milp_imb}-(d).

For completeness up to $n_R$ stages, observe that Stage~II explicitly enumerates
$k=\bar{s},\bar{s}+1,\dots,n_R$ and, for each $k$, solves \eqref{eq:milp_imb} with $k$
stages as a feasibility-and-optimization problem. If \eqref{eq:milp_imb} is feasible
for some $k^\star\le n_R$, then when the loop reaches $k^\star$ the solver will return
a feasible solution, and the algorithm terminates with $\bar{s}\le k^\star$.
\end{proof}

The procedure is sound with respect to the staged formulation \eqref{eq:milp_imb}.
Moreover, it is complete up to $n_R$ stages in the sense of Prop.~\ref{prop:completeness}:
if a staged solution exists for some $k\le n_R$, the algorithm will find one by
enumerating the number of stages.


\emph{Solution complexity:} We address the NP-hardness of the underlying ILP formulations by exploiting structural properties that allow most decision variables to be handled through LP relaxations. While LPs admit polynomial-time algorithms in theory, the practical efficiency of the proposed approach stems from the elimination of integer variables at the motion-planning level, rather than from worst-case complexity guarantees.
In Stage~I, the LP relaxation~\eqref{eq:milp}--LP involves $n_P + |T| + n_\mathcal{Y} + 1$ decision variables and $2\cdot n_P + 2\cdot n_\mathcal{Y} + n_d$ linear constraints. In Stage~II, the size of the refinement problem~\eqref{eq:milp_imb} depends on the number of synchronization stages~$\bar{s}$. Specifically, the formulation contains $\bar{s}(n_P+|T|)+n_\mathcal{Y}$ variables and $2 \cdot \bar{s} \cdot n_P + 2\cdot n_\mathcal{Y} + n_d$ constraints. As shown in the simulations, $\bar{s}$ typically remains small, which keeps the overall problem size manageable even for large robotic teams.

\emph{Limitations:} One of the limitations of the method proposed is the interpretation of the Boolean goal which indicates the position of the robots in their final state and not along paths. Moreover, narrow corridors may lead to an increased number of synchronizations, which will slow down the execution as increased coordination will be required. Another limitation arises from the potentially large number of binary variables of vector $\b{x}$, corresponding to a high number of regions of interest. This can increase the computational load while solving the MILP formulation in the second stage of the method, but despite these constraints, the algorithm remains highly efficient for a wide range of practical scenarios.

\section{Simulations}\label{sec:6}

This section presents simulation results across several scenarios, demonstrating both the efficiency and scalability of the proposed path-planning algorithm. In all experiments, the regions of interest and the initial robot positions are randomly generated. The simulations were executed on a workstation equipped with  an AMD Ryzen 9 9950X 16 CPU and 64 GB of RAM. The MATLAB implementation,  available at \url{https://github.com/IoanaHustiu/efficient_TAPF_for_Boolean_specifications.git}, employs  the \emph{intlinprog} solver for efficient handling of the optimization constraints.

\subsection{Task-Assignment and Path Finding (TAPF) problem}

\begin{figure}[ht]
    \centering
    \captionsetup{justification=centering}
    \includegraphics[width=.8\linewidth]{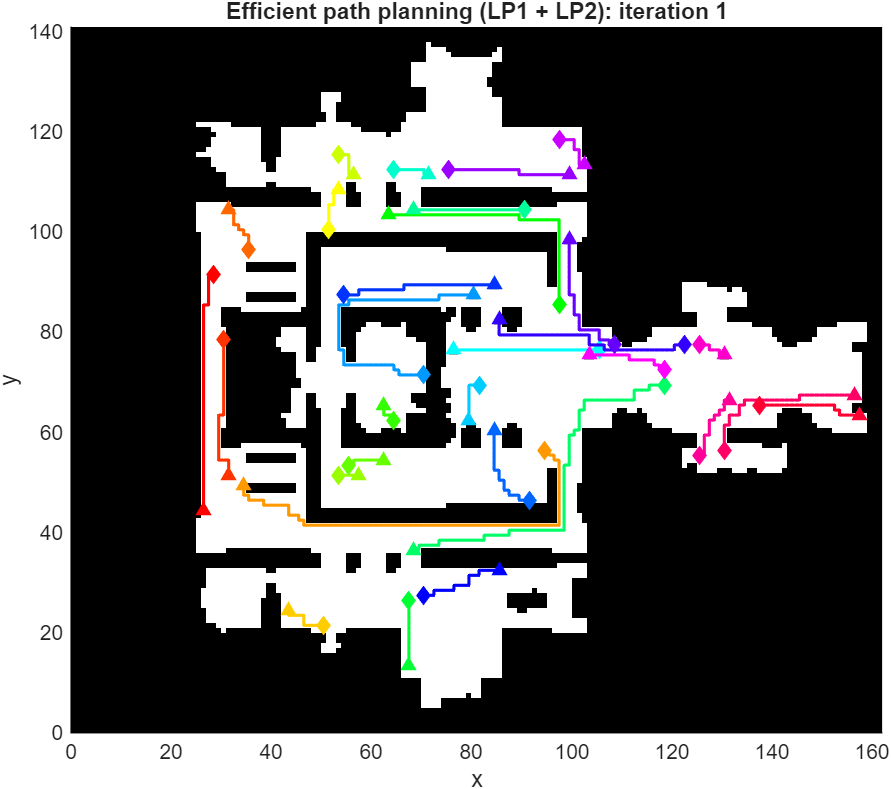}
\caption{Example of the \textit{ht\_chantry} benchmark environment used for TAPF simulations. 
Black pixels denote obstacles, while colored markers represent robots and their assigned goal regions for a trial involving 30 robots. }
\label{fig:TAPF_30robots}
\end{figure}

We evaluate the proposed algorithm on the \textit{ht\_chantry} benchmark map from \cite{stern2019multi}. A pixel-level abstraction is adopted, where \emph{each free pixel corresponds to a place} in the RMPN and transitions represent 4-neighborhood adjacency. The original image has a resolution of $141\times162$ pixels; after removing obstacles (black pixels in Fig.~\ref{fig:TAPF_30robots}, which illustrates a trial with 30 robots), the environment comprises $7461$ free pixels (places) and $27926$ transitions (adjacency arcs). The complete MATLAB implementation is publicly available as script \texttt{main\_TAPF\_chantry.m}.

For each team size, 20 random start--goal configurations were generated from the benchmark map, and the planner optimizes the total traveled distance (or number of movements) for task assignment and collision-free movement. We apply Algorithm~\ref{alg:two_stage} in a TAPF special case where $\b{x}=\mathbf{1}$ (a conjunctive global goal over $n_R$ disjoint goal regions). Thus, $\b{x}$ can be omitted from the optimization.

\begin{itemize}
\item \textbf{Stage I}: Solve LP \eqref{eq:milp_m}; if the resulting trajectories are non-intersecting (i.e., $s^*=1$), the process stops.
\item \textbf{Stage II (if needed)}: If intersections occur, solve LP \eqref{eq:milp_mc}, a particular case of MILP \eqref{eq:milp_imb} with $\b{x}=\mathbf{1}$, introducing intermediate markings to ensure collision-free motion.
\end{itemize}

\begin{table}[!t]
\centering
\caption{\small TAPF problem: mean values for proposed algorithm vs. ILP formulation}
\label{tab:reachability}
{\scriptsize 
\begin{tabular}{|c|cc|c|ccc|c|}
\hline
\multirow{2}{*}{$n_R$} &
  \multicolumn{2}{c|}{runtime (s) (\eqref{eq:milp_m} + \eqref{eq:milp_mc})} &
  \multirow{2}{*}{cost} &
  \multicolumn{3}{c|}{synchronizations $\bar{s}$} &
  \multirow{2}{*}{SR \%} \\
& ours (LPs) & ILPs & & mean & min & max & \\ \hline
10   & 0.39  & 0.44  &  450.7 & 1.00 & 1 & 1 & 100 \\
50   & 1.40  & 2.37  & 1329.1 & 1.40 & 1 & 2 & 100 \\
100  & 4.75  & 5.85  & 2029.3 & 2.00 & 1 & 3 & 100 \\
250  & 6.30  & 8.47  & 3003.0 & 2.65 & 2 & 5 & 100 \\
500  & 9.98  & 15.83 & 4413.2 & 3.52 & 2 & 5 & 85  \\
750  & 12.77 & 20.83 & 5320.5 & 4.06 & 2 & 5 & 80  \\
1000 & 13.15 & 31.87 & 5794.3 & 4.50 & 4 & 5 & 50  \\
1250 & 12.50 & 20.48 & 6300.1 & 4.77 & 4 & 5 & 45  \\
1500 & 13.08 & 23.49 & 6462.5 & 5.00 & 5 & 5 & 10  \\
1750 & 10.34 & 38.61 & 6724.5 & 5.00 & 5 & 5 & 10  \\
2000 & 3.83  & 37.99 & 5837.0 & 4.50 & 4 & 5 & 10  \\
2250 & 3.52  & 22.37 & 7009.0 & 5.00 & 5 & 5 & 5   \\
2500 & --    & --    & --      & --   & --& --& 0   \\
\hline
\end{tabular}
} 
\end{table}

Table~\ref{tab:reachability} summarizes, for each $n_R$: (i) \emph{runtime\_LP} = runtime of our approach (LP \eqref{eq:milp_m} plus, when necessary, LP \eqref{eq:milp_mc}); (ii) \emph{runtime\_ILP} = runtime when solving \eqref{eq:milp_m} and \eqref{eq:milp_mc} as \emph{integer} programs (a baseline needed); (iii) \emph{cost} $=\sum_{i=1}^{\bar{s}} \mathbf{1}^\top \cdot \b{\sigma}_i$ (total number of movements); (iv) $\bar{s}_{\text{mean}}$, $\bar{s}_{\min}$, $\bar{s}_{\max}$ = statistics of the required synchronizations; and (v) \emph{SR\_percent} = success rate (\%). Dividing \emph{cost} by $n_R$ yields the \emph{average path length per robot}, which grows moderately with $n_R$ as expected.

The main limitation arises from memory usage when multiple synchronizations are required, since the number of variables in LP~\eqref{eq:milp_mc} grows proportionally to $\bar{s}(|P|+|T|) $. When $\bar{s}>5$, the constraint matrix may exceed available memory, explaining the observed drop in success rate. In practice, this issue can be mitigated by solving a simplified MILP version with $s$ as the only integer variable - allowing minor local adjustments for collisions - or by coarsening the map resolution, which reduces the number of places and transitions and keeps the problem tractable for larger teams.

\subsection{Boolean-Based Specifications}
\label{subsec:boolean_warehouse}

\begin{figure*}[!t]
    \centering
    \includegraphics[width=.7\textwidth]{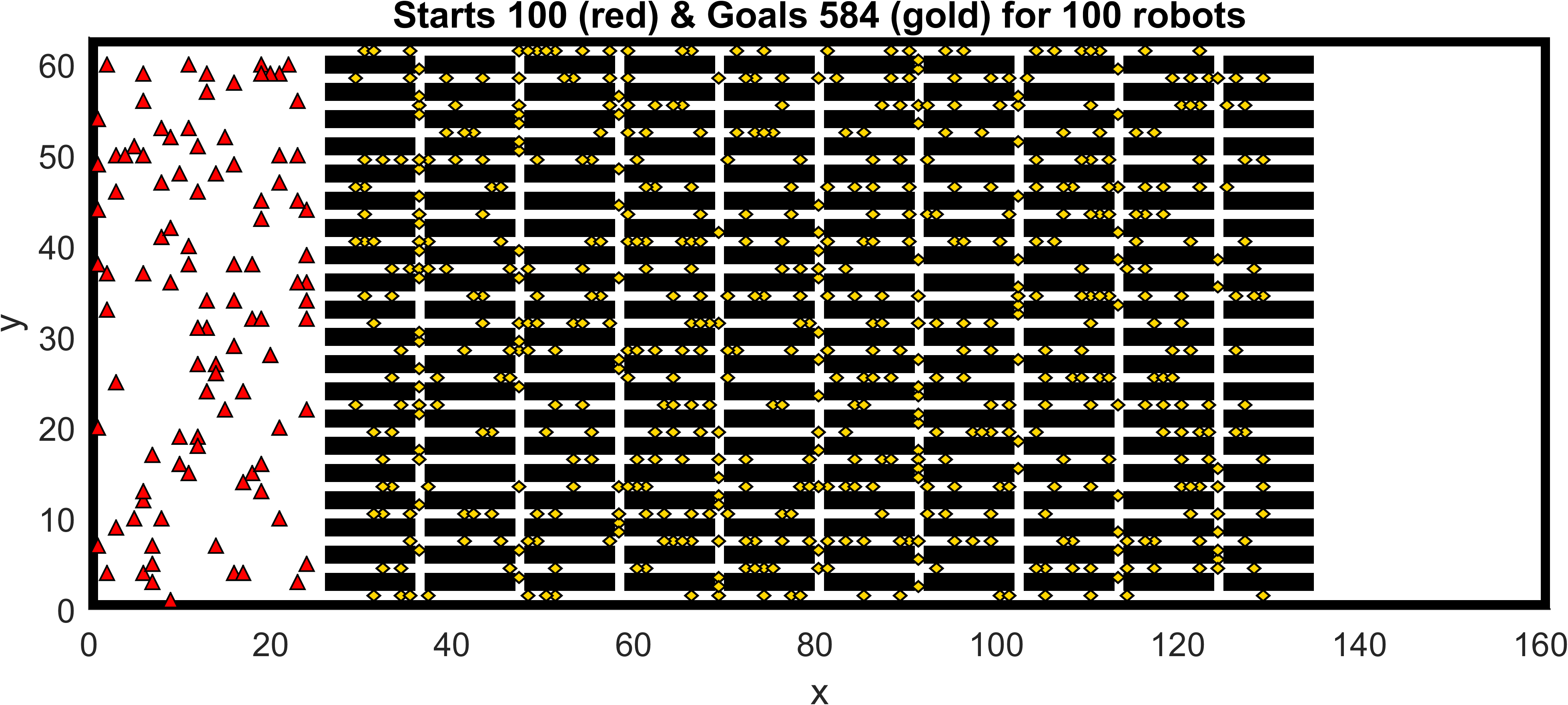}
    \caption{Example of warehouse environment used in Boolean-based experiments.
    Robots start on the left side and must reach central regions that satisfy a randomly generated Boolean formula. Each corridor can host at most one robot at a time.}
    \label{fig:boolean_warehouse}
\end{figure*}

In this set of experiments, we evaluate the performance of the proposed method when the global task is expressed as a Boolean formula containing multiple disjunctions per term. We consider a fixed team of 100 robots operating in the \textit{warehouse} environment (Fig.~\ref{fig:boolean_warehouse})\cite{stern2019multi}, which consists of narrow corridors of unitary width, allowing only one robot to traverse a corridor at a time. The warehouse layout includes 21 entrances to the shelf corridors, each permitting the passage of a single robot simultaneously. Consequently, the number of intermediate markings required for synchronization is given by the ceiling of the ratio between the number of robots and the number of available corridor entrances, i.e., $\lceil n_R / 21 \rceil$. The robots are initially placed randomly on the left side of the map and must reach a set of target regions located in the central area of the warehouse, thereby fulfilling the Boolean goal.

The Boolean specification is generated as a conjunction of $n_R$ terms, where each term is a disjunction of a randomly chosen number of destination regions:
\[
\varphi = (y_1^1 \vee \dots \vee y_{k_1}^1) \wedge (y_1^2 \vee \dots \vee y_{k_2}^2) \wedge \dots \wedge (y_1^{n_R} \vee \dots \vee y_{k_{n_R}}^{n_R}),
\]
with each $k_i$ randomly selected such that $1 \le k_i \le n_P$. The parameter $n_P$ means the maximum number of terms in a disjunction and it is varied between 1 and 10, where $n_P=1$ corresponds to the TAPF case (each robot having a single destination). For each configuration, 20 independent trials are performed, with both the initial positions and goal regions generated randomly in each trial. The MATLAB implementation is given as script \texttt{main\_TAPF\_aisle\_boolean.m}.

\begin{table}[!t]
\centering\footnotesize
\caption{Boolean-based specification: mean values for 100 robots in the warehouse environment}
\label{tab:boolean_warehouse}
\begin{adjustbox}{width=.49\textwidth}
\begin{tabular}{|c|cc|c|c|c|}
\hline
$n_P$ & runtime \eqref{eq:milp_imb}-MILP & runtime \eqref{eq:milp_imb}-ILP \cite{kloetzer2020path} & cost & $\bar{s}$ & SR(\%) \\
\hline
1  & 19.35 & 57.51 & 6911.1 & 5 & 100 \\
2  & 12.38 & 48.24 & 6050.4 & 5 & 100 \\
3  & 13.22 & 55.93 & 5477.6  & 5 & 100 \\
4  & 14.96 & 48.04 & 5068.9 & 5 & 100 \\
5  & 14.59 & 51.3 & 4775.9  & 5 & 100 \\
6  & 16.01 & 48.31 & 4562 & 5 & 100 \\
7  & 14.74 & 46.63 & 4255.4 & 5 & 100 \\
8  & 15.38 & 50.97 & 4229.7 & 5 & 100 \\
9  & 13.71 & 45.73 & 3977.6 & 5 & 100 \\
10 & 15.45 & 50.75 & 3856.3 & 5 & 100 \\
\hline
\end{tabular}
\end{adjustbox}
\end{table}

Table~\ref{tab:boolean_warehouse} reports the mean values over the experiments for both our the current approach and the ILP formulation from \cite{kloetzer2020path}. The results show that the computational time remains nearly constant as $n_P$ increases. This behavior indicates that the problem complexity is primarily determined by the number of intermediate markings ($\bar{s}$) required for synchronization, which, as explained earlier, remains constant in this setup, rather than the number of disjunctions in the formula. Although adding disjunctions increases the number of possible choices for each robot, the overall computational effort of solving MILP~\eqref{eq:milp} and MILP~\eqref{eq:milp_imb} remains almost constant as long as $\bar{s}$ does not change.

It should be noted that, in this case, the presence of a Boolean formula introduces binary decision variables $\b{x}$ into the optimization, corresponding to the possible destinations for the robots. Consequently, the problem becomes a MILP, where the number of binary variables is equal to the number of regions included in the Boolean specification.

Additional experiments, including large-scale evaluations and various strategies for rounding the assignment vector $\b{x}$ in the MILP formulation~\eqref{eq:milp_imb}, are reported in~\cite{IPHuKlMa25}, where several LP-based rounding techniques are compared in terms of runtime and solution quality across diverse robotic scenarios. These results further validate the efficiency and robustness of the proposed PN-based formulation for solving complex Boolean-goal planning problems.

\begin{figure*}[t]
    \centering
    \begin{subfigure}[t]{0.245\textwidth}
        \centering
        \includegraphics[width=\linewidth]{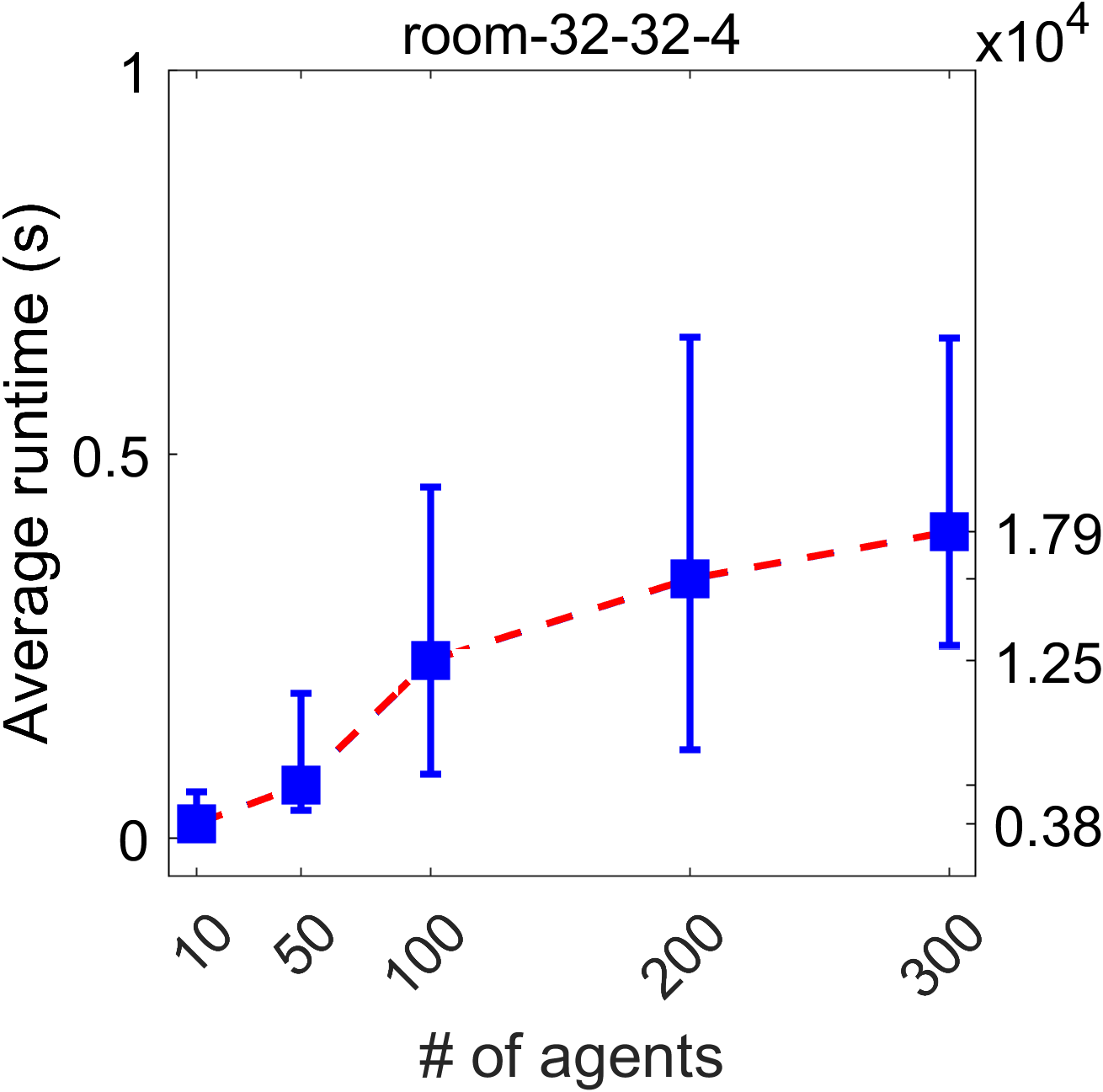}
        \subcaption{}
        \label{fig:rt_room}
    \end{subfigure}\hfill
    \begin{subfigure}[t]{0.2275\textwidth}
        \centering
        \includegraphics[width=\linewidth]{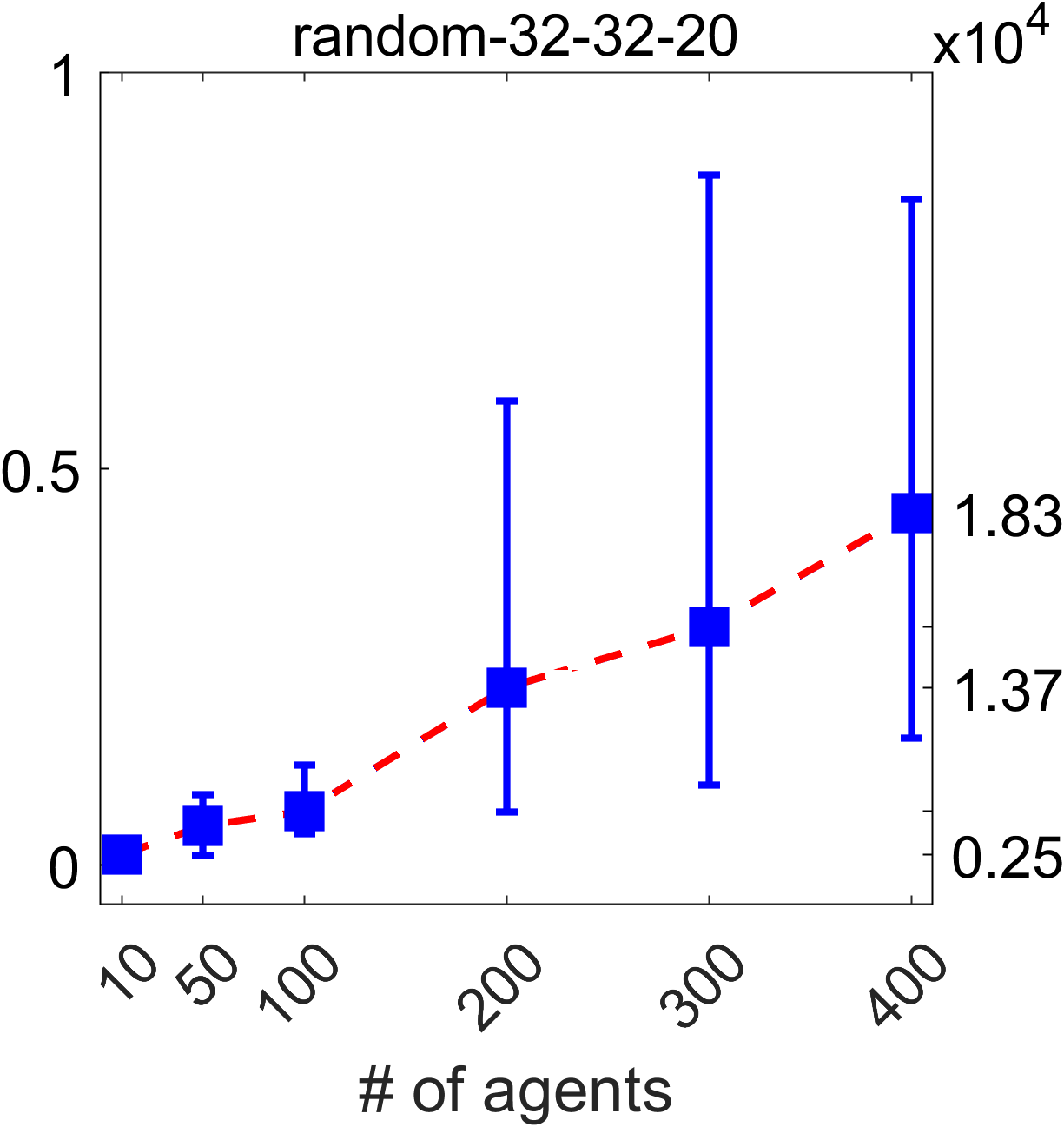}
        \subcaption{}
        \label{fig:rt_random}
    \end{subfigure}\hfill
    \begin{subfigure}[t]{0.2275\textwidth}
        \centering
        \includegraphics[width=\linewidth]{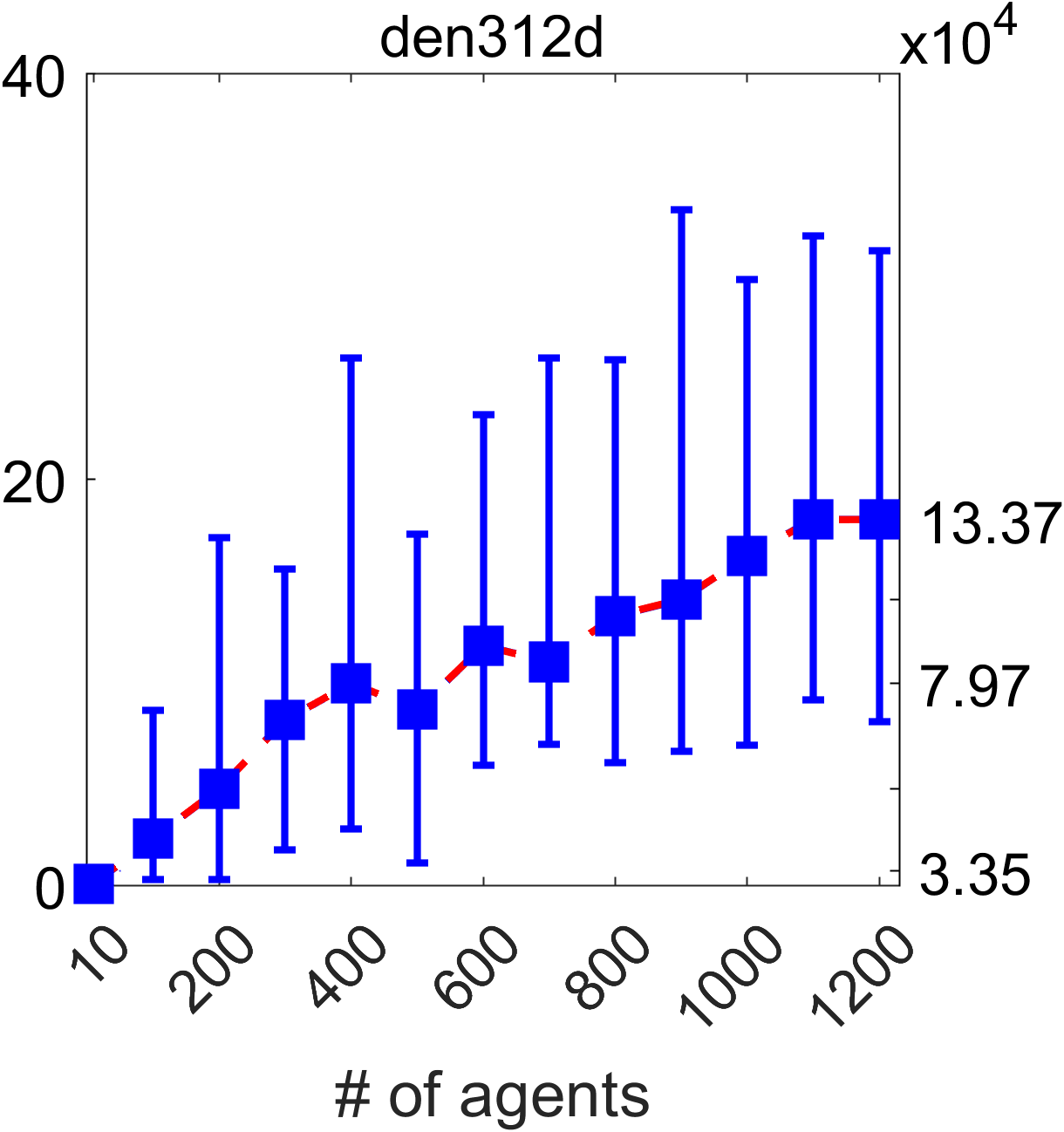}
        \subcaption{}
        \label{fig:rt_den312d}
    \end{subfigure}\hfill
    \begin{subfigure}[t]{0.245\textwidth}
        \centering
        \includegraphics[width=\linewidth]{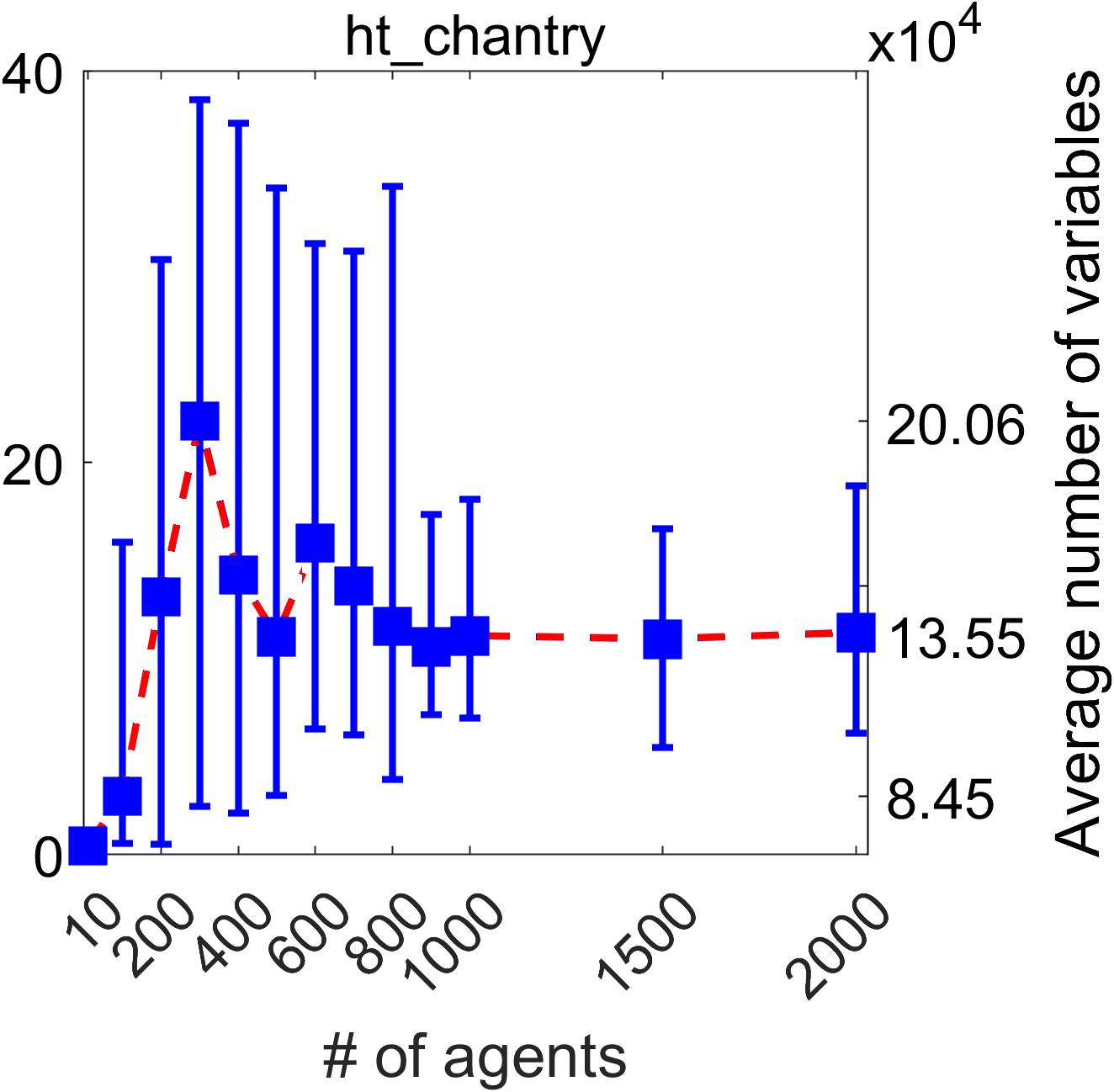}
        \subcaption{}
        \label{fig:rt_ht}
    \end{subfigure}
    \caption{Runtime measured in seconds for \textit{room-32-32-4, random-32-32-20, den312d}, and \textit{ht\_chantry} maps.}
    \label{fig:runtime_benchmarks}
\end{figure*}

\subsection{Comparative Evaluation with MAPF Baselines}

To further contextualize our contributions, we evaluated the proposed method on four standard MAPF benchmarks from~\cite{stern2019multi}: \textit{room-32-32-4} ($32{\times}32$, 682 states), \textit{random-32-32-20} ($32{\times}32$, 819 states), \textit{den312d} ($65{\times}81$, 2445 states), and \textit{ht\_chantry} ($162{\times}141$, 7461 free cells). These maps are widely used to assess scalability in MAPF research, particularly for Conflict-Based Search (CBS)~\cite{sharon2015conflict} and Priority-Based Search (PBS)~\cite{friedrich2024scalable,ma2019searching}, yet no existing TAPF approaches have been tested on such large instances.

For each map, 25 random scenarios were generated with the number of robots scaled to roughly half of the available cells, ensuring a unit cell capacity (one robot per cell). Problem~\eqref{eq:milp_mc} was solved using its LP relaxation, imposing solver runtime upper-bounds of 30\,s for the smaller maps and 60\,s for the larger ones. Figure~\ref{fig:runtime_benchmarks} reports the mean, minimum, and maximum runtimes versus team size, along with the corresponding number of decision variables.

Across all benchmarks, our method achieved a 100\% success rate and near-linear growth in runtime, demonstrating strong computational scalability. Compared with classical MAPF baselines, it remains competitive or faster: Enhanced and Explicit Estimation CBS~\cite{li2021eecbs} show rapid performance degradation beyond 100 agents, while the MCPP algorithm~\cite{friedrich2024scalable} maintains full success up to 3000 robots but under less expressive formulations. Similarly, MAPF-LNS2~\cite{li2022mapf} requires roughly an order of magnitude longer runtimes on the same maps. Recent large-scale studies on \textit{ht\_chantry}~\cite{ren2024dms,pertzovsky5013125adapting,veerapaneni2025windowed,ali2024improved,gandotra2025anytime} reach 1000 robots with specialized heuristics; in contrast, our PN-based approach attains comparable scalability while additionally handling Boolean goal specifications and offering formal guarantees through total unimodularity.

Overall, the proposed formulation bridges the gap between optimization-based TAPF and heuristic MAPF solvers, combining the tractability of LP relaxations with the expressiveness of logical task modeling. It thus provides a unified and scalable framework for large-team coordination where both optimality and logical correctness are required.

\section{Conclusions}\label{sec:7}

This paper presents a novel approach for obtaining efficiently collision-free path planning together with task allocation in the context of large teams of mobile robots that have the goal to fulfill a Boolean-based specification. Our primary contribution is to uncover and demonstrate the unique structural properties of the formulated optimization problem - the total unimodularity of the constraint matrix - which allows us to avoid solving the classical ILP formulation and use its MILP and LP relaxations. Moreover, the achieved results through the theoretical component are supported by multiple performed simulations with a balance between efficiency and accuracy, making the approach highly effective for applications with teams of around 2000 robots. Furthermore, the method guarantees optimality with respect to the congestion-aware cost function induced by the synchronization-on-demand formulation, even though the problem formulation requires introducing new constraints. Future work is aiming to reduce the limitation regarding the expressivity of the global task, first by managing visiting regions along paths and next by employing Temporal Logics such as LTL.

\bibliographystyle{IEEEtran}
\bibliography{REF.bib}

\begin{IEEEbiography}[{\includegraphics[width=1in,height=1.25in,clip,keepaspectratio]{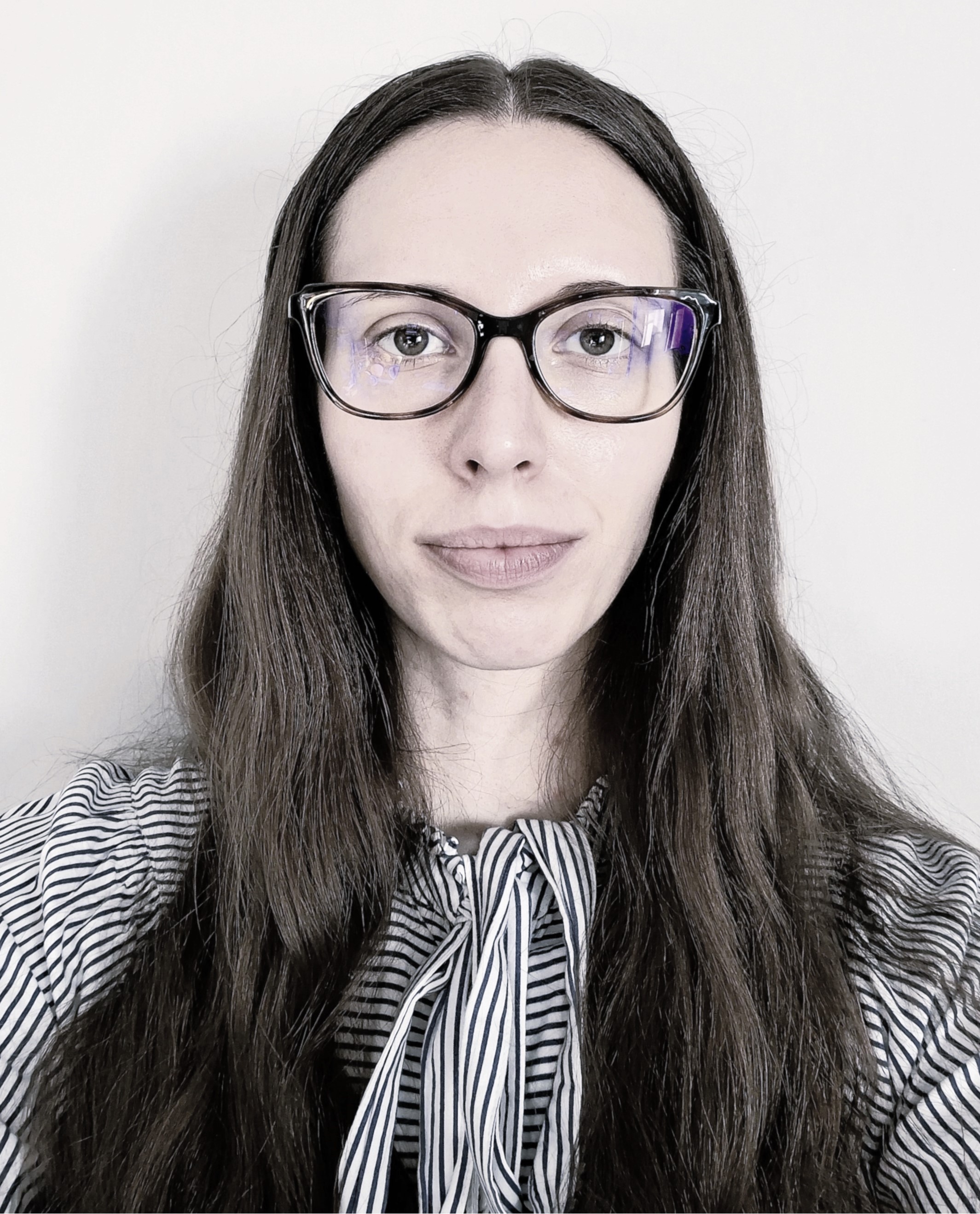}}]{Ioana Hustiu} received the B.S. in 2020 and the M.Sc. degree in 2022 in automatic control and applied informatics from the Technical University of Iasi, Romania, where she is a Ph.D. student.

Her research interests include task allocation and path planning in context of distributing high-level specification for multi-robot systems using discrete event systems.
\end{IEEEbiography}

\begin{IEEEbiography}[{\includegraphics[width=1in,height=1.25in,clip,keepaspectratio]{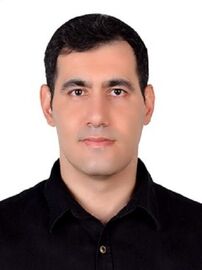}}]{Roozbeh Abolpour} received the B.Sc. degrees, one in electrical engineering control and another in computer engineering from Shiraz University, Shiraz, Iran, in 2012 and 2014, respectively, the M.Sc. degree in electrical engineering with a concentration in control from the Sharif University of Technology, Tehran, Iran, in 2014, and the Ph.D. degree in electrical engineering, specializing in control, from Shiraz University, Shiraz, Iran, in 2020. 

He is currently a Postdoctoral Researcher with the Energy Information Networks and Systems Group at the Technical University of Darmstadt, Darmstadt, Germany. His research interests include control systems, data-driven model predictive control, quadratically constrained quadratic programming, and the optimal power flow problem.
\end{IEEEbiography}

\begin{IEEEbiography}[{\includegraphics[width=1in,height=1.25in,clip,keepaspectratio]{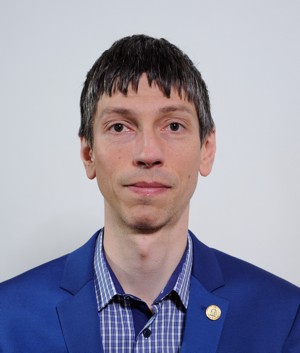}}]{Marius Kloetzer} received the B.S. and M.Sc. degrees in computer science from the Technical University of Iasi, Romania, in 2002 and 2003, respectively, and the Ph.D. degree in systems engineering from Boston University, MA, USA, in 2008. He is currently a Full Professor with the Technical University of Iasi, Romania. His research interests include formal tools for discrete event systems with applications in motion planning for mobile robots.

Marius Kloetzer was a visiting researcher at Ghent University, Belgium, and at the University of Zaragoza, Spain. He has been Organizing Committee chair at ICSTCC'2017 and Work-in-Progress co-chair at ETFA'2019.
\end{IEEEbiography}

\begin{IEEEbiography}[{\includegraphics[width=1in,height=1.25in,clip,keepaspectratio]{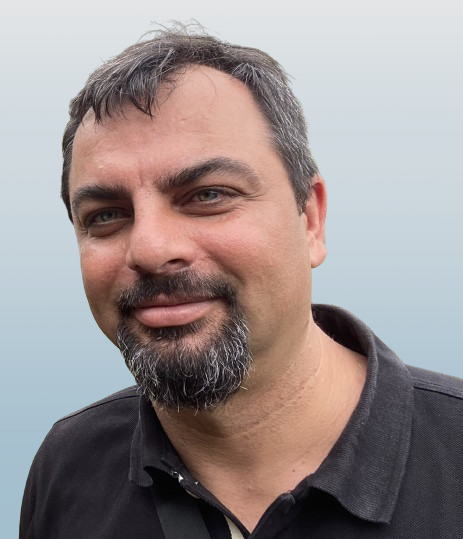}}]{Cristian Mahulea} received his B.S. and M.Sc. degrees in control engineering from the Technical University of Iasi, Romania, in 2001 and 2002, respectively, and his Ph.D. in systems engineering from the University of Zaragoza, Spain, in 2007. Currently, he is a Full Professor at the University of Zaragoza, where he chaired the Department of Computer Science and Systems Engineering from 2020 to 2024. He has also served as a visiting professor at the University of Cagliari, Italy, and has been a visiting researcher at the University of Sheffield (UK), Boston University (USA), University of Cagliari (Italy), and ENS Paris-Saclay (France).

He is currently an Associate Editor for IEEE Transactions on Automatic Control (TAC), the International Journal of Robotics Research (IJRR), and Discrete Event Dynamic Systems: Theory and Applications (JDES). He was the General Chair of ETFA 2019 and was AE for IEEE Trans. on Automation Science and Engineering (TASE) and IEEE Control Systems Letters (L-CSS).
\end{IEEEbiography}

\appendices

\section{Proof of Theorem~\ref{th:TU_final_m}}\label{ap:TU1}

We prove total unimodularity of the matrix
\[
\b{A}
\;=\;
\begin{bmatrix}
\b{C}\\
\b{Post}
\end{bmatrix}
\in \{-1,0,1\}^{2|P|\times |T|}
\]
using the Ghouila--Houri characterization (Theorem~\ref{th:GH}).

Since the RMPN is a \emph{state machine}, each transition
$t\in T$ has exactly one input place and one output place, i.e., $
|\preset{t}| = |\postset{t}| = 1.$

Let $p_{\preset{t}}$ and $p_{\postset{t}}$ denote the unique places in
$\preset{t}$ and $\postset{t}$, respectively.
By construction of the RMPN, the $t$-th column of $\b{C}$ and $\b{Post}$ satisfies
\[
\b{C}[p_{\preset{t}},t]=-1,\qquad
\b{C}[p_{\postset{t}},t]=+1,\qquad
\b{Post}[p_{\postset{t}},t]=1,
\]
and all other entries in that column are zero.

We index the rows of $\b{A}$ as follows.
Rows $1,\dots,|P|$ correspond to the places of $\b{C}$, and rows
$|P|+1,\dots,2|P|$ correspond to the places of $\b{Post}$.
For each place $p\in P$, let
\[
r_C(p)=p,\qquad r_P(p)=|P|+p.
\]

Let $R\subseteq\{1,\dots,2|P|\}$ be an arbitrary subset of rows.
We construct a partition $R=R_1\cup R_2$ that satisfies the Ghouila--Houri condition. Let
\begin{align*}
R_1 &:= \bigl(R\cap\{1,\dots,|P|\}\bigr)
      \;\cup\;
      \{\,r_P(p)\in R \mid r_C(p)\notin R\,\},\\
R_2 &:= R\setminus R_1.
\end{align*}
That is, all selected rows of $\b{C}$ are placed in $R_1$.
A selected row $r_P(p)$ of $\b{Post}$ is placed in $R_2$ if and only if the
corresponding row $r_C(p)$ of $\b{C}$ is also selected.

Column-wise verification. Fix an arbitrary column $t\in T$.
As noted above, the only nonzero entries in column $t$ are
$$
\b{A}[r_C(p_{\preset{t}}),t]=-1,
\b{A}[r_C(p_{\postset{t}}),t]=+1,
\b{A}[r_P(p_{\postset{t}}),t]=+1.
$$

We examine their contributions under the partition $(R_1,R_2)$.

\emph{Output place $p_{\postset{t}}$.} There are four possibilities:
\begin{itemize}
\item If both $r_C(p_{\postset{t}})\in R$ and $r_P(p_{\postset{t}})\in R$, then
$r_C(p_{\postset{t}})\in R_1$ and $r_P(p_{\postset{t}})\in R_2$.
The signed contributions $+1$ and $-1$ cancel, yielding $0$.
\item If $r_C(p_{\postset{t}})\in R$ and $r_P(p_{\postset{t}})\notin R$, the signed contribution is $+1$.
\item If $r_C(p_{\postset{t}})\notin R$ and $r_P(p_{\postset{t}})\in R$, then
$r_P(p_{\postset{t}})\in R_1$, and the signed contribution is $+1$.
\item If neither row is selected, the contribution is $0$.
\end{itemize}
Thus, the combined signed contribution of the two output-place rows belongs to $\{0,1\}$.

\emph{Input place $p_{\preset{t}}$.} If $r_C(p_{\preset{t}})\in R$, then it belongs to $R_1$ and contributes $-1$.
Otherwise, it contributes $0$.

Summing the input and output contributions, we obtain
\[
\sum_{r\in R_1} \b{A}[r,t] - \sum_{r\in R_2} \b{A}[r,t]
\;\in\;
\{0,1\} + \{0,-1\}
\;=\;
\{-1,0,1\}.
\]

Since $R$ and $t$ were arbitrary, the Ghouila--Houri criterion holds for $\b{A}$.
Therefore, $\b{A}$ is totally unimodular. \hfill $\blacksquare$

\section{Proof of Theorem~\ref{th:inter}}\label{ap:TU2}

We prove that the constraint matrix of \eqref{eq:milp_mc} is TU
using the Ghouila--Houri criterion.
Since TU is preserved under row permutations and multiplication of rows by $-1$,
we equivalently represent the stage-coupling equalities as
\[
\b{m}_{i-1} + \b{C}\cdot \b{\sigma}_i - \b{m}_i = \b{0},\qquad i=1,\dots,\bar{s},
\]
and keep the stage-wise unit-capacity constraints as
\[
\b{Post}\cdot \b{\sigma}_i + \b{m}_{i-1} \le \b{1},\qquad i=1,\dots,\bar{s}.
\]
This is identical to \eqref{eq:milp_mc} up to row sign changes and therefore has the same TU property. Let the decision variables be
\[
\b{z} = (\b{m}_0,\b{m}_1,\dots,\b{m}_{\bar{s}},\b{\sigma}_1,\dots,\b{\sigma}_{\bar{s}}),
\]
where $\b{m}_0$ is fixed and may be moved to the right-hand side without affecting TU.
Index the rows by stage $i=1,\dots,\bar{s}$ and place $p\in P$ as follows: (i) equality rows $\mathcal{E}_i(p)$ encoding $\b{m}_{i-1}[p] + (\b{C}\cdot \b{\sigma}_i)[p] - \b{m}_i[p]=0$; (ii) capacity rows $\mathcal{K}_i(p)$ encoding
$\b{m}_{i-1}[p] + (\b{Post}\cdot \b{\sigma}_i)[p]\le 1$.

Fix an arbitrary subset of rows $R$.
Define a partition $R=R_1\cup R_2$ by:
(i) every selected equality row belongs to $R_1$; and
(ii) a selected capacity row $\mathcal{K}_i(p)$ belongs to $R_2$ if and only if
$\mathcal{E}_i(p)\in R$, and otherwise belongs to $R_1$.

We now verify the Ghouila--Houri condition column-wise.

\medskip
\noindent\emph{Firing variables $\b{\sigma}_i[t]$.}
Fix a stage $i$ and transition $t\in T$.
Since the RMPN is a state machine, $|\preset{t}|=|\postset{t}|=1$; let
$p_{\preset{t}}$ and $p_{\postset{t}}$ denote the unique input and output places.
The column of $\b{\sigma}_i[t]$ has nonzeros only in
$\mathcal{E}_i(p_{\preset{t}})$, $\mathcal{E}_i(p_{\postset{t}})$ and
$\mathcal{K}_i(p_{\postset{t}})$, with coefficients $-1$, $+1$, and $+1$.
By the above partition rule, the signed contributions of
$\mathcal{E}_i(p_{\postset{t}})$ and $\mathcal{K}_i(p_{\postset{t}})$ cancel when both
rows are selected and otherwise sum to $+1$; adding the possible $-1$ from
$\mathcal{E}_i(p_{\preset{t}})$ yields a total in $\{-1,0,1\}$.

\medskip
\noindent\emph{Marking variables $\b{m}_j[p]$.}
Fix $p\in P$ and $j\in\{0,\dots,\bar{s}\}$.
The variable $\b{m}_j[p]$ appears only in the rows
$\mathcal{E}_j(p)$ (coefficient $-1$ if $j\ge 1$),
$\mathcal{E}_{j+1}(p)$ (coefficient $+1$ if $j\le\bar{s}-1$),
and $\mathcal{K}_{j+1}(p)$ (coefficient $+1$ if $j\le\bar{s}-1$).
If $\mathcal{E}_{j+1}(p)\in R$, then $\mathcal{K}_{j+1}(p)\in R_2$ and the
corresponding signed contributions cancel; otherwise any selected
$\mathcal{K}_{j+1}(p)$ lies in $R_1$.
In all cases the signed sum belongs to $\{-1,0,1\}$.

\medskip
Hence, for every row subset $R$ and every column, the Ghouila--Houri condition holds,
and the constraint matrix of \eqref{eq:milp_mc} is TU. \hfill $\blacksquare$

\end{document}